\documentclass[10pt,twocolumn,letterpaper]{article}

\usepackage{iccv}
\usepackage{times}
\usepackage{epsfig}
\usepackage{graphicx}
\usepackage{amsmath}
\usepackage{amssymb}
\usepackage{multirow}
\usepackage{wasysym}
\usepackage{stmaryrd}
\usepackage{times}
\usepackage{booktabs}
\usepackage{subcaption}
\usepackage{float}

\newcommand{\refsec}[1]{Section~\ref{sec:#1}}

\newcommand{\reffig}[1]{Figure~\ref{fig:#1}}
\newcommand{\reftab}[1]{Table~\ref{tab:#1}}

\iccvfinalcopy % *** Uncomment this line for the final submission

\def\iccvPaperID{26} % *** Enter the ICCV Paper ID here

\ificcvfinal\fi
\begin{document}

\definecolor{ceiling}{RGB}{0,255,0}
\definecolor{floor}{RGB}{0,0,255}
\definecolor{wall}{RGB}{0,255,255}
\definecolor{beam}{RGB}{255,255,0}
\definecolor{column}{RGB}{255,0,255}
\definecolor{window}{RGB}{100,100,255}
\definecolor{door}{RGB}{200,200,100}
\definecolor{table}{RGB}{170,120,200}
\definecolor{chair}{RGB}{255,0,0}
\definecolor{sofa}{RGB}{200,100,100}
\definecolor{bookcase}{RGB}{10,200,100}
\definecolor{board}{RGB}{200,200,200}
\definecolor{clutter}{RGB}{50,50,50}

\definecolor{Terrain}{RGB}{200, 90, 0}
\definecolor{Tree}{RGB}{0, 128, 50}
\definecolor{Vegetation}{RGB}{0, 220, 0}
\definecolor{Building}{RGB}{255, 0, 0}
\definecolor{Road}{RGB}{100, 100, 100}
\definecolor{GuardRail}{RGB}{200, 200, 200}
\definecolor{TrafficSign}{RGB}{255, 0, 255}
\definecolor{TrafficLight}{RGB}{255, 255, 0}
\definecolor{Pole}{RGB}{128, 0, 255}
\definecolor{Misc}{RGB}{255, 200, 150}
\definecolor{Truck}{RGB}{0, 128, 255}
\definecolor{Car}{RGB}{0, 200, 255}
\definecolor{Van}{RGB}{255, 128, 0}

\definecolor{Unknown_}{RGB}{0, 0, 0}
\definecolor{Grass_}{RGB}{0, 204, 0}
\definecolor{Ground_}{RGB}{77, 128, 0}
\definecolor{Pavement_}{RGB}{179, 204, 230}
\definecolor{Hedge_}{RGB}{128, 128, 0}
\definecolor{Topiary_}{RGB}{0, 179, 179}
\definecolor{Rose_}{RGB}{230, 0, 0}
\definecolor{Obstacle_}{RGB}{51, 51, 230}
\definecolor{Tree_}{RGB}{77, 179, 26}
\definecolor{Background_}{RGB}{26, 26, 26}
%%%%%%%%% TITLE
\title{Exploring Spatial Context for 3D Semantic Segmentation of Point Clouds}

\author{Francis Engelmann\textsuperscript{\textdagger},
Theodora Kontogianni\textsuperscript{\textdagger},
Alexander Hermans and Bastian Leibe\\
Computer Vision Group, Visual Computing Institute\\ RWTH Aachen University\\
{\tt\small \{engelmann,kontogianni,hermans,leibe\}@vision.rwth-aachen.de}
% For a paper whose authors are all at the same institution,
% omit the following lines up until the closing ``}''.
% Additional authors and addresses can be added with ``\and'',
% just like the second author.
% To save space, use either the email address or home page, not both
% \and
% Second Author\\
% Institution2\\
% First line of institution2 address\\
% {\tt\small secondauthor@i2.org}
}

\maketitle

\renewcommand{\thefootnote}{\fnsymbol{footnote}}
\footnotetext[2]{Both authors contributed equally. Order decided by coin flip.}

\begin{abstract}
Deep learning approaches have made tremendous progress in the field of semantic segmentation over the past few years.
However, most current approaches operate in the 2D image space. Direct semantic segmentation of unstructured 3D point clouds is still an open research problem.
The recently proposed PointNet architecture presents an interesting step ahead in that it can operate on unstructured point clouds, achieving encouraging segmentation results.
However, it subdivides the input points into a grid of blocks and processes each such block individually.
In this paper, we investigate the question how such an architecture can be extended to incorporate larger-scale spatial context.
We build upon PointNet and propose two extensions that enlarge the receptive field over the 3D scene.
We evaluate the proposed strategies on challenging indoor and outdoor datasets and show improved results in both scenarios.
\end{abstract}

\section{Introduction}
Semantic segmentation is an important capability for intelligent vehicles, such as autonomous cars or mobile robots.
Identifying the semantic meaning of the observed 3D structure around the vehicle is a prerequisite for solving subsequent tasks such as navigation or reconstruction \cite{EngelmannGCPR16_shapepriors, EngelmannWACV17_samp}.
Consequently, the problem has attracted a lot of attention, and notable successes have been achieved with the help of deep learning techniques.
However, most state-of-the-art semantic segmentation approaches operate on 2D images, which naturally lend themselves to processing with Convolutional Neural Networks (CNNs) \cite{LongCVPR2015,ChenPK0Y16,WuSH16a,Noh_2015_ICCV}.

Processing unstructured 3D point clouds, such as those obtained from LiDAR or stereo sensors, is a much harder problem, and it is only recently that first successful deep learning approaches have been proposed for this task \cite{voxnet,HuangICPR2016,YiSGG2016,pointnet}.
Such point clouds can be obtained from LiDAR sensors mounted on top of a recording vehicle or they can be obtained from visual SLAM approaches operating on the vehicle's cameras \cite{Kasyanov2017_VISLAM}.
Finding approaches that can directly operate on point cloud data is highly desirable, since it avoids costly preprocessing and format conversion steps. However, the question what is the best network architecture to process unstructured 3D point clouds is still largely open.

\begin{figure}[t!]
	\centering
  \includegraphics[width=0.5\textwidth, trim=0 30 0 0, clip]{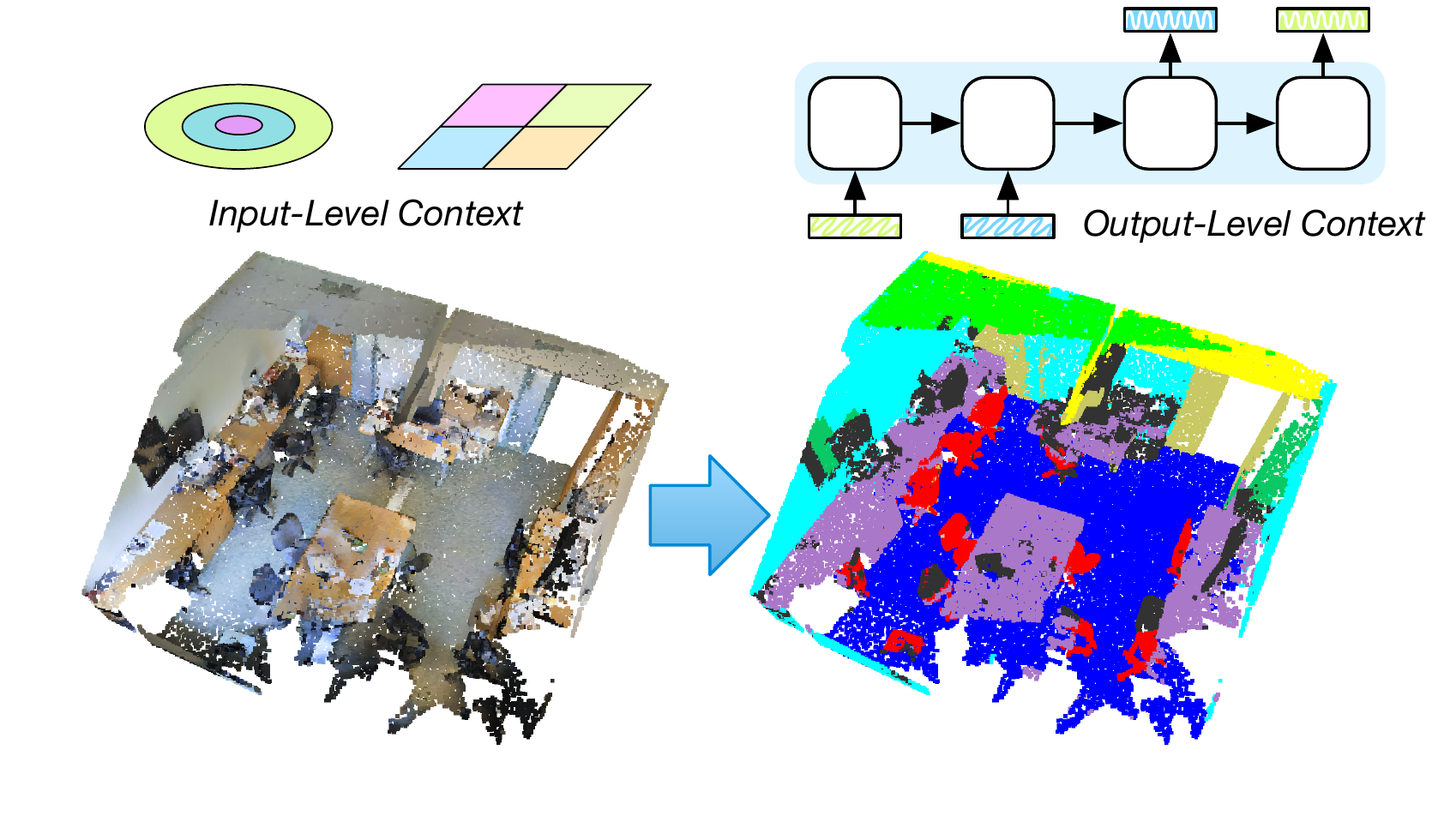}
	\caption{We explore mechanisms to extend the spatial context for 3D semantic segmentation of point clouds.}
	\label{fig1}
	\vspace{-10px}
\end{figure}

In this paper, we take inspiration from the recent PointNet work by Qi \etal \cite{pointnet}, which currently defines the state of the art in 3D semantic segmentation.
PointNet learns a higher dimensional spatial feature representation for each 3D point and then aggregates all the points within a small 3D volume (typically an occupancy grid cell) in order to bring in some form of 3D neighborhood context. However, this neighborhood context is very restricted, as each grid cell is processed independently.

In this paper, we investigate possible mechanisms to incorporate context into a point cloud processing architecture. We focus on spatial context, which has been identified as being very important for semantic segmentation \cite{MottaghiCVPR2014, Pohlen2017CVPR}. We introduce two mechanisms to add spatial context to an existing PointNet. The first mechanism incorporates neighborhood information by processing input data from multiple scales or multiple adjacent regions together (input-level context). The second mechanism operates on the estimated point descriptors and aims at consolidating them by exchanging information over a larger spatial neighborhood (output-level context). For both mechanisms, we explore several possible realizations and compare them experimentally. As our results show, both mechanisms improve semantic segmentation quality.

\textbf{Contributions.} The key contributions of our work can be summarized as follows:
(1) We present two mechanisms that can be used to incorporate spatial context into semantic 3D point cloud segmentation.
(2) We show how these mechanisms can be incorporated into the PointNet pipeline.
(3) We verify experimentally that our proposed extensions achieve improved results on challenging indoor and outdoor datasets.

\section{Related Work}
\textbf{Unstructured Point Clouds.} A varied number of sensors and setups exist which help to obtain unstructured point clouds: areal data from airborne laser scanners, laser scanners mounted on dynamic setups in a push-broom configuration \cite{robotcar}, rotating lasers e.g. Velodyne \cite{kitti}, or static lasers \cite{semantic3d.net}.
Additionally, indoor spaces can be scanned using devices such as the Microsoft Kinect \cite{Silberman:ECCV12} or Matterport cameras \cite{s3dis}.
All these devices produce point clouds of different quality and density.
We apply our method to indoor data from \cite{s3dis} and to synthetic urban outdoor data from \cite{vkitti}.

\textbf{Traditional Methods.}
Hackel et al. \cite{hackel2016fast} use traditional random forest classifiers with 3D features (without color). Their method is based on eigenvalues and eigenvectors of covariance tensors created by the nearest neighbors of the points. Their main contribution is an efficient approximate nearest neighbors computation at different scales.
Munoz et al. \cite{Munoz_3DPVT_2008} follow a similar approach but replace the random forest classifier with an associative Markov network.
Random forest classifiers are also used in \cite{ZhangICRA2015} to classify data from 2D images and 3D point clouds, which they later fuse. Similarly, Xu et al. \cite{XuMVA2013} fuse camera and LiDAR sensor data.
Xiong et al. \cite{Xiong_ICRA_2011} propose a sequential parsing procedure that learns the spatial relationships of objects. Lai et al. \cite{LaiICRA14} introduce a hierarchical sparse coding technique for learning features from synthetic data.
Vosselman et al. \cite{vosselman2013point} combine multiple segmentation and post-processing methods to achieve useful point cloud segmentations.

\textbf{Deep-learning Methods.}
In a deep learning context, point clouds can be represented in a regular volumetric grid in order to apply 3D convolutions \cite{voxnet,HuangICPR2016}.
Alternatively, 3D points can be mapped to a 2D representation followed by 2D convolutions \cite{vmv3d}.
In \cite{3dor}, the authors are performing 2D convolutions in 2D snapshots of a 3D point cloud and then project the labels back to 3D space.
In \cite{OndruskaRSS2016} a deep learning framework learns semantic segmentation by tracking point clouds.
Yi et al. \cite{YiSGG2016} use spectral CNNs on 3D models represented as shape graphs for shape part segmentation.
Recent methods operate directly on raw point clouds with KD-trees \cite{kdnets} or fully convolutional layers \cite{pointnet}.

\section{Method}
In this section we start by reviewing the PointNet model, then we introduce our mechanisms of extending context and finish by describing our two exemplary architectures.

\subsection{PointNet}
\begin{figure}[t]
	\centering
	\includegraphics[width=1.0\linewidth]{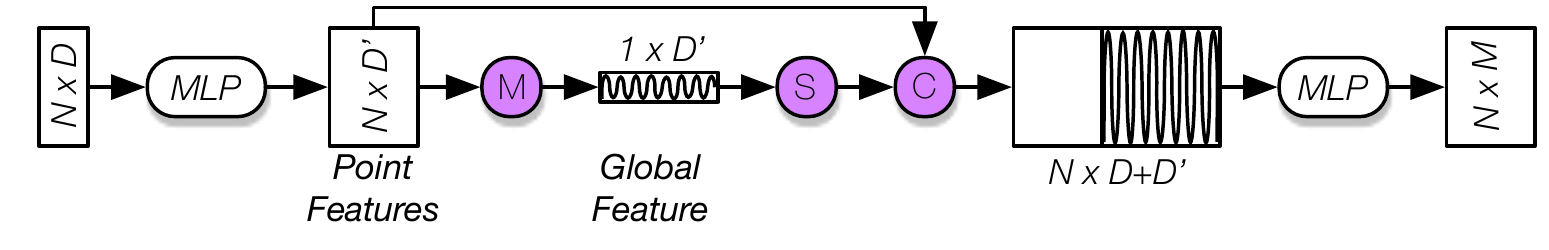} 	
	\caption{\textbf{Simplified PointNet Architecture.}
	In this work, we build upon the PointNet architecture for semantic segmentation.
	In short, it computes a global feature which summarizes a set of input points.
	Specifically, the network takes $N$ points as input, applies a series of multi-layer-perceptrons transformations and aggregates the point features by max pooling them into a global feature.
	Global and local features are concatenated and the per point class scores are returned.
	 (MLP): Multi-Layer-Perception, (M): Max-Pool, (S): Vertical Stack, (C): Concatenate. See text and Qi et al. \cite{pointnet} for more details.}
	\label{fig:pointnet_model}
\vspace{-10px}
\end{figure}

PointNet \cite{pointnet} is a deep neural network that, when used for semantic segmentation, takes as input a point cloud and outputs the per point semantic class labels.
First, it splits a point cloud into 3D blocks, then it takes $N$ points inside a block and after a series of Multi-Layer-Perceptrons (MLP) per point, the points are mapped into a higher dimensional space $D'$, these are called local \emph{point-features}.
Max-pooling is applied to aggregate information from all the points resulting in a common \emph{global-feature} invariant to input permutations.
The global-feature is then concatenated with all the point-features. After another series of MLPs these combined features are used to predict the $M$ output class scores.
\reffig{pointnet_model} shows a simplified model.

\textbf{Caveats.} The global-features in PointNet summarize the context of a single block (block-feature), as a result the aggregated information is passed only among points inside the same block. \\
\\
Context outside a block is equally important and could help make more informed class label predictions.
Therefore we introduce two mechanisms to add context: \textbf{input-level context} -- which operates directly on the input point clouds -- and \textbf{output-level context}  -- which consolidates the output from the input-level context.

\begin{figure*}[ht!]
	\centering
	\includegraphics[width=1.0\linewidth]{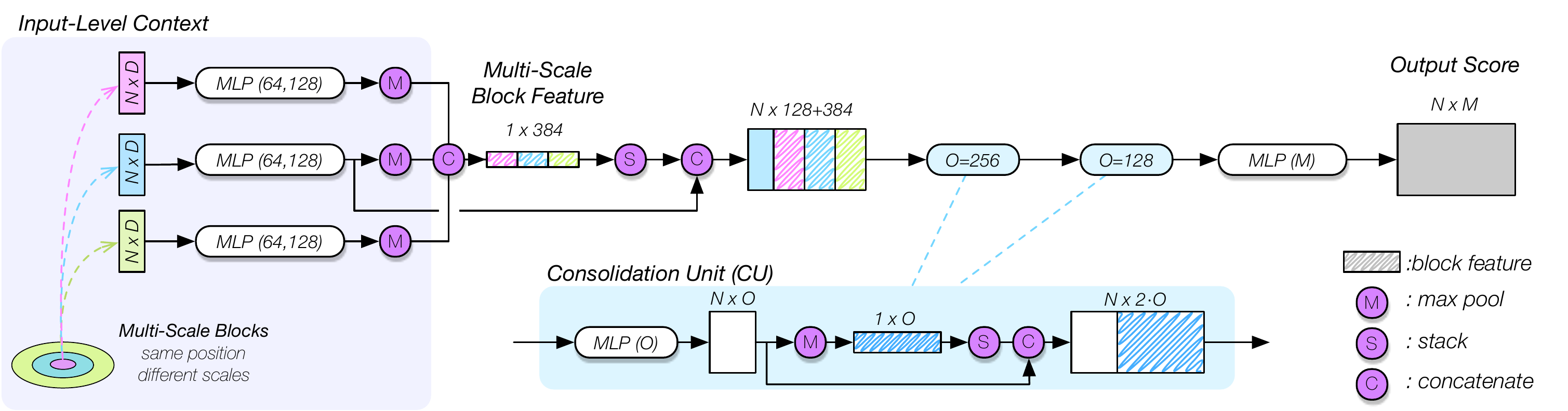} 
	\caption{\textbf{Architecture with multi-scale input blocks and consolidation units (MS-CU).} %
	The network takes as input three blocks from multiple scales, each one containing $N$ $D$-dimensional points.
	Separately, for each scale, it learns a block-feature similarly to the PointNet mechanism.
	The concatenated block-features are appended to the input-features and then transformed by a sequence of consolidation units (see \refsec{consolidations}).
	The network outputs per point scores.
	Shaded fields represent block-features.}
	\label{fig:scale_model}
	\vspace{-15px}
\end{figure*}
\subsection{Input-Level Context}
\label{sec:input_level_context}
\begin{figure*}[t!]
	\centering
	\includegraphics[width=1.0\linewidth]{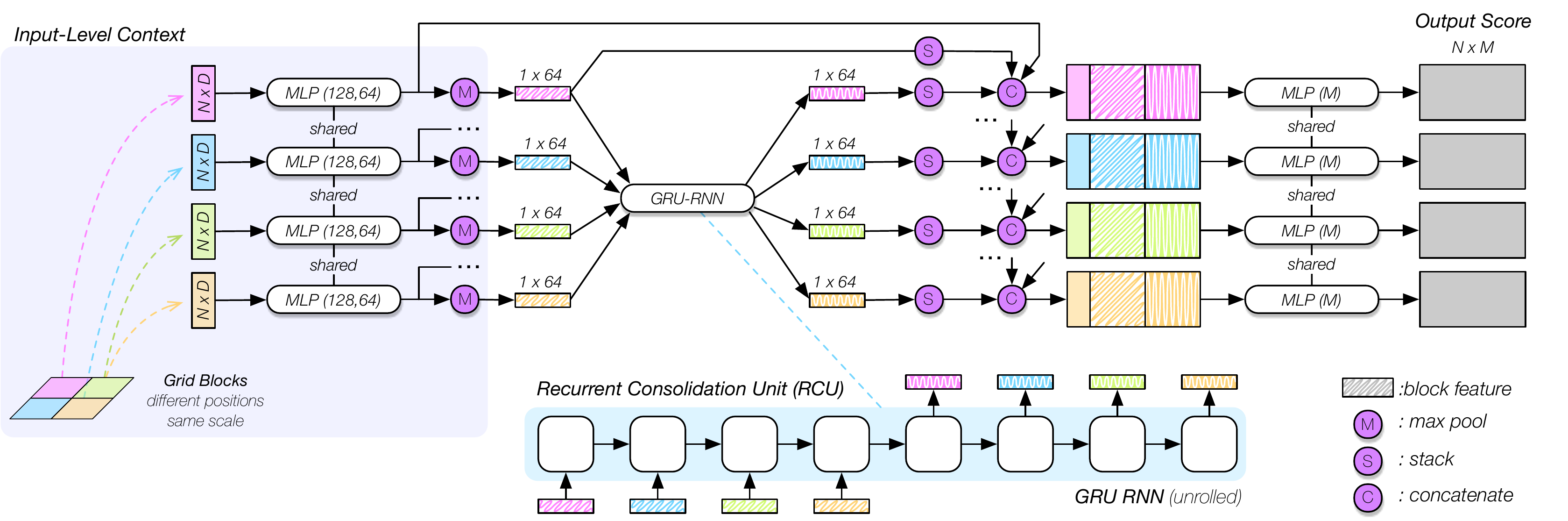} 
	\caption{\textbf{Architecture with grid input blocks and a recurrent consolidation unit (GB-RCU).}
	The network takes as input four blocks from a grid structure,
	each one containing $N$ $D$-dimensional points.
	It then learns the block-features using the same MLP weights for each block.
	All block-features are passed through a recurrent consolidation unit (see \refsec{consolidations}) which shares the spatial context among all blocks and returns updated block-features. 
	The updated block-features are appended to the input-features together with the original block-features and used to compute the output per point scores.
	Shaded fields represent block-features. 
	Some skip-connections are omitted for clarity.}
	\label{fig:trans_model}
\end{figure*}

In this straightforward addition, we increase the context of the network by considering a group of blocks simultaneously instead of one individual block at a time as done in PointNet. Context is shared among all blocks in a group.
These groups of blocks are selected either from the same position but at multiple different scales (\textbf{Multi-Scale Blocks}, see \reffig{scale_model}, left) or from neighboring cells in a regular grid (\textbf{Grid Blocks}, see \reffig{trans_model}, left).
For each input block, we compute a block-feature using the mechanism from PointNet.
For the multi-scale version, we train a block-descriptor for each scale individually to obtain scale-dependent block-features.
In the case of grid blocks, all block features are computed by a shared single-scale block-descriptor.
In the end, both approaches output a set of block-features corresponding to the input blocks.

\subsection{Output-Level Context}
\label{sec:consolidations}
At this stage, we further consolidate the block-features obtained from the previous stage.
Here, we differ between two consolidation approaches:
\\
\textbf{Consolidation Units (CU)} consume a set of point features, transform them into a higher dimensional space using MLPs and apply max-pooling to generate a common block-feature which is again concatenated with each of the high dimensional input features (see \reffig{scale_model}, blue box).
This procedure is similar to the block-feature mechanism of PointNet.
The key point is that CUs can be chained together into a sequence CUs forming a deeper network.
The intuition behind this setup is as follows:
In the beginning each point sees only its own features.
After appending the block-features, each point is additionally informed about the features of its neighboring points.
By applying CUs multiple times, this shared knowledge is reinforced.
\\ \\
\textbf{Recurrent Consolidation Units (RCU)} are the second type of context consolidation we employ.
RCUs take as input a sequence of block-features originating from spatially nearby blocks and return a sequence of corresponding updated block-features.
The core idea is to create block-features that take into consideration neighboring blocks as well.
In more detail, RCUs are implemented as RNNs, specifically GRUs \cite{gru}, which are a simpler variation of standard LSTMs \cite{lstm}.
GRUs have the capability to learn long range dependencies. That range can either be over time (as in speech recognition) or over space as in our case.
The cells of the unrolled GRU are connected in an unsynchronized  many-to-many fashion (see \reffig{trans_model}, blue box).
This means that the updated block-features are returned only after the GRU has seen the whole input sequence of block-features.
Intuitively, GRU retain relevant information about the scene in their internal memory and update it according to new observations.
We use this memory mechanism to consolidate and share the information across all input blocks.
For example, the decision about whether a point belongs to a chair is changed if the network remembers that it has seen a table further down in the room.

In the following, we describe two exemplary architectures which combine the previously introduced components.
For those, we provide a detailed evaluation and report improved results in  \refsec{experiments}.

\subsection{Multi-Scale (MS) Architecture}
\label{sec:MS}
The full MS architecture is displayed in \reffig{scale_model}.
The learned block-features from the multi-scale blocks, (see \refsec{input_level_context}) are concatenated into one \textbf{multi-scale block-feature}.
This multi-scale block-feature is further concatenated with the transformed input point-features and passed through a series of CUs (see \refsec{consolidations}).
Applying a final MLP results in output scores for each input point.

Specific for this architecture is the sampling procedure to select the positions of the multi-scale blocks:
We randomly pick a $D$-dimensional point from the input point cloud as the center of the blocks and we group together $N$ randomly selected points that fall within a specified radius.
This procedure is repeated at the same point for multiple radii.

\subsection{Grid (G) Architecture}
\reffig{trans_model} shows the pipeline of the architecture with grid input blocks. It consists of the following components:
The input level context is a group of four blocks from a 2x2 grid-neighborhood (see \refsec{input_level_context}) is fed into
a series of MLPs that transform the point features, with weights shared among all blocks.
These block-features are passed to an RCU that updates the individual block-features with common context from all neighboring blocks. The updated block-features are then concatenated with the original block-features.
They are then used, along with the local features, for class predictions.
After a series of fully connected layers the output of class scores is computed for each point.

\section{Experiments}
\label{sec:experiments}

For experimental evaluation, we compare our two architectures with PointNet \cite{pointnet}, the current state-of-the-art semantic segmentation method directly operating on point clouds.
We produce quantitative results for our models and the baseline on two challenging datasets:
\emph{Stanford Large-Scale 3D Indoor Spaces} (S3DIS) \cite{s3dis} and on \emph{Virtual KITTI} (vKITTI) \cite{vkitti}.
Additionally, we provide qualitative results on point clouds obtained from a Velodyne HDL-64E LiDAR scanner from the \emph{KITTI} dataset \cite{kitti}.
We will now describe these datasets in more detail.

\textbf{Stanford Large Scale 3D Indoor Scenes.}
This dataset is composed of 6 different large scale indoor areas, mainly conference rooms, personal offices and open spaces.
It contains dense 3D point clouds scanned using a \emph{Matterport} camera. Each point is labeled with one of the 13 semantic classes listed in \reftab{s3dis_color_iou}.
Using the reference implementation of PointNet, we were able to reproduce the results reported by Qi et al. \cite{pointnet}, see \reftab{s3dis_color}.
Throughout the paper, we follow the same evaluation protocol used in \cite{pointnet}, which is a 6-fold cross validation over all the areas.

\begin{figure}[b]
	\centering
\includegraphics[width=0.24\textwidth, trim=100 260 200 0, clip]{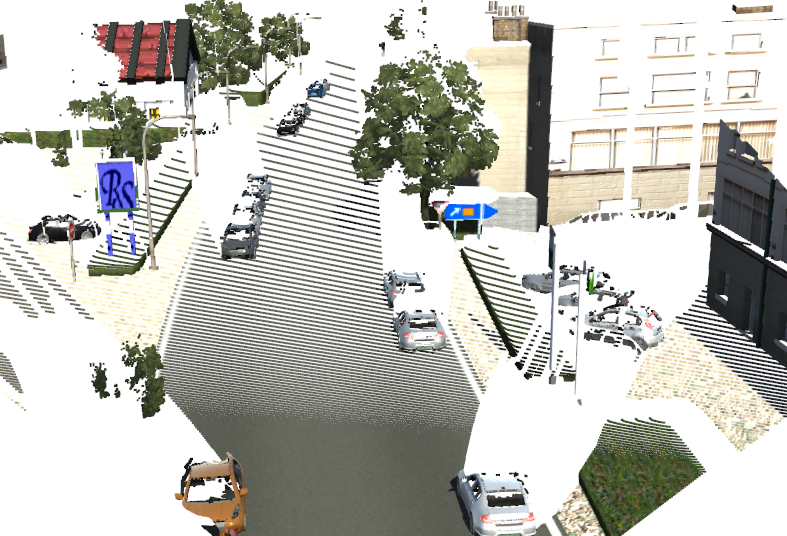}%
\includegraphics[width=0.24\textwidth, trim=0 0 0 150, clip]{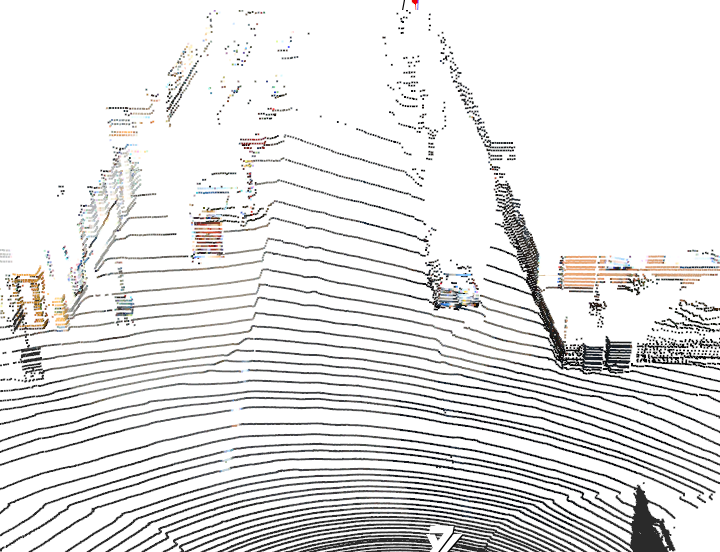}%
	\caption{We train our network on synthetic point clouds generated from vKITTI \cite{vkitti} (left) and apply it onto real-world Velodyne LiDAR  point clouds (right). The structure and the varying density are comparable.}
	\label{fig:pc_density}
\end{figure}

\textbf{Virtual KITTI.}
Due to the lack of semantically annotated large-scale outdoor datasets, we rely on the photo-realistic synthetic vKITTI dataset which closely mimics the real-world KITTI dataset.
It consists of 5 different monocular video sequences in urban settings, fully annotated with depth and pixel-level semantic labels.
In total there are 13 semantic class, listed in \reftab{s3dis_iou}.
For our purposes, we project the given 2D depth into 3D space to obtain semantically annotated 3D point clouds.
Conveniently, this procedure results in point clouds that resemble the varying density of real world point clouds obtained by Velodyne LiDAR scanners (see \reffig{pc_density}).
For test and training, we split the original sequences into 6 non-overlapping subsequences.
The final train-test sets are created by choosing point clouds from each subsequence at regular time-intervals.
For evaluation, we also follow the 6-fold cross validation protocol.

\subsection{Evaluation Measures}
As in \cite{pointnet}, we evaluate on the intersection over union (IoU), the average per class accuracy and overall accuracy.
Intersection over union is computed as:
\begin{equation}
IoU = \frac{TP}{TP+FP+FN}
\end{equation}
where TP is the number of true positives, FP the number of false positives and FN the number of false negatives.

\subsection{Quantitative Results}
In this section, we analyze the effectiveness of the input-block schemes and the consolidation units exemplary on the two previously introduced models. As input features, we differentiate between geometry (XYZ) and geometry with color (XYZ+RGB).

\textbf{Geometry with Color.}
First, we compare the grid-blocks in combination with a recurrent consolidation block (G+RCU) to the original PointNet.
Using the same evaluation setup as described in \cite{pointnet} we are able to show improved results over PointNet, see \reftab{s3dis_color} and \reftab{s3dis_color_iou}.
This proves our hypothesis that RCU are able to convey context among blocks and thus improving results. 
During training, each room is split into blocks of 1x1 m on the ground plane. Each block extends over the whole room height. Neighboring blocks are overlapping by 0.5 meters in both directions. We select four blocks simultaneously from a 2x2 grid-neighborhood (see \reffig{trans_model}, left). Each block contains 4096 points.
The unrolled GRU is 8 cells long (4 input, 4 output). It's memory size is 64. During testing, the room is split into non-overlapping blocks and evaluated on all 2x2 groups of blocks. Each block is evaluated only once.

Next, we take a look at the multi-scale input block with consolidation units (MS-CU) model.
To build the multi-scale blocks, we follow the process described in  \refsec{MS}. As radii, we choose [0.25, 0.5, 1.0] m. As distance metric
we choose the \emph{Chebyshev}-distance which generates axis-aligned rectangular blocks. The middle scale block is equal to the PointNet block regarding shape and size.

By using sampling (necessary for the multi-scale block construction), we diverge from the previous training procedure so we re-run all experiments under these new conditions.

We validate the influence of each of the architecture's components by adding them one-by-one to our pipeline and evaluating after each step, see \reftab{s3dis_color} and \reftab{s3dis_color_iou}.
First, we only consider the input-level context i.e the multi-scale block feature (MS) as input to our pipeline while skipping the consolidation units.
This shows some performance benefits over PointNet but not as much as one would expect considering the enlarged input context. Next, we take only single-scale input blocks and add one consolidation unit (SS+CU(1)). The results show that the CU outperforms the MS input blocks. It also shows that CUs provide a simple technique to boost the network's performance. Finally, we combine both the MS blocks and the CU while appending another CU to the network (MS+CC(2)). This full model is depicted in \reffig{scale_model}.

\textbf{Geometry only.}
Until now, each input point was described by a 9-dimensional feature vector $[X,Y,Z,R,G,B,X',Y',Z']$
where $[X,Y,Z]$ are the spatial coordinates of a point, $[R,G,B]$ its color and $[X',Y',Z']$ the normalized coordinated based on the size of the environment, see \cite{pointnet} for further details.
Without doubt, color is a very strong input feature in the context of semantic segmentation.
In this section, we pose the question what will happen if no color information is available like it is the case with point clouds obtained from laser scanners.
To simulate the missing colors, we simply discard the color information from the input feature and re-run the experiments. \reftab{s3dis} and \ref{tab:s3dis_iou} show the obtained results. See caption for discussion of the results.

\subsection{Qualitative Results}
\begin{figure}[t]
\centering
\includegraphics[width=0.45\linewidth]{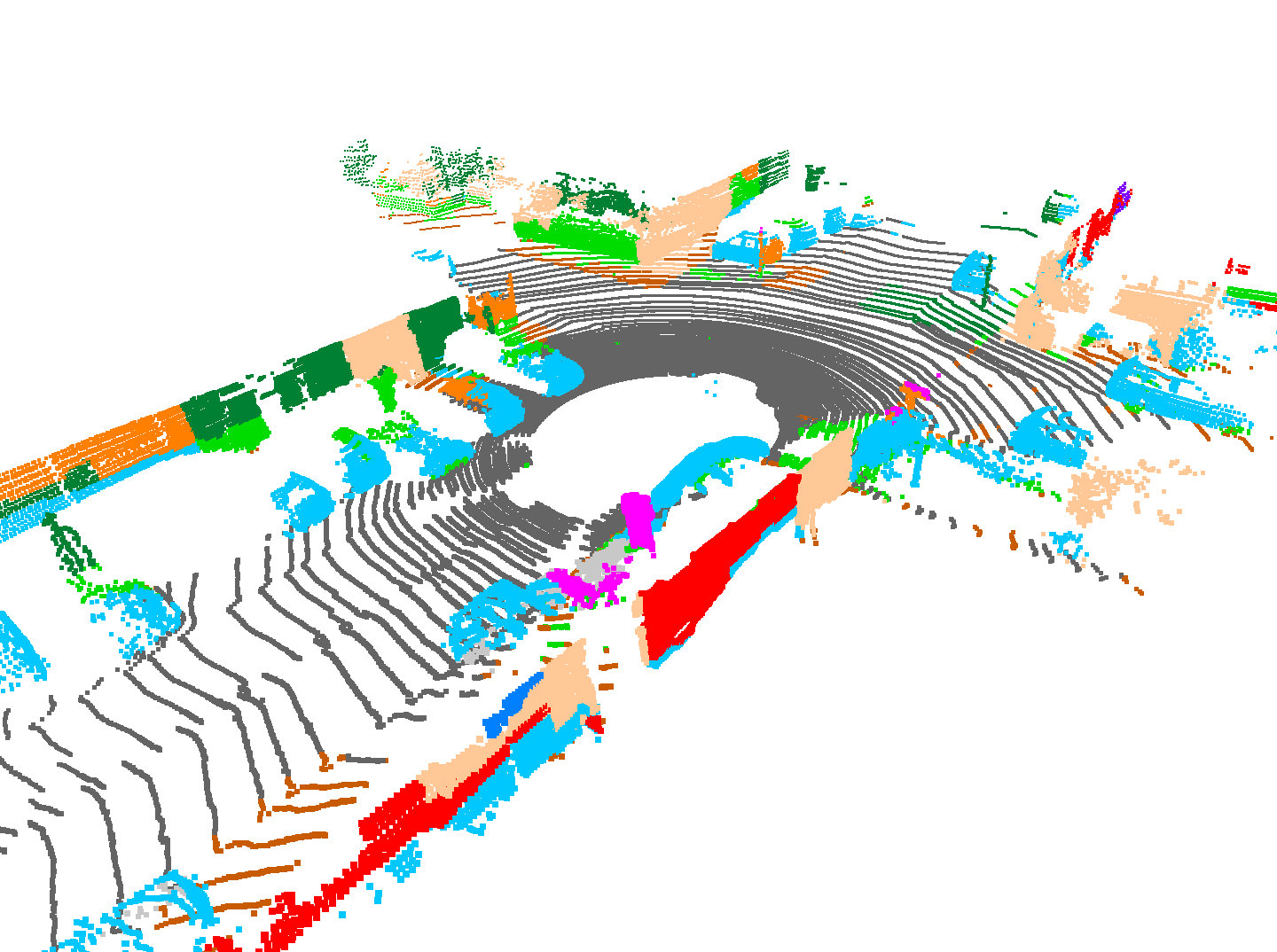}%
\includegraphics[width=0.45\linewidth]{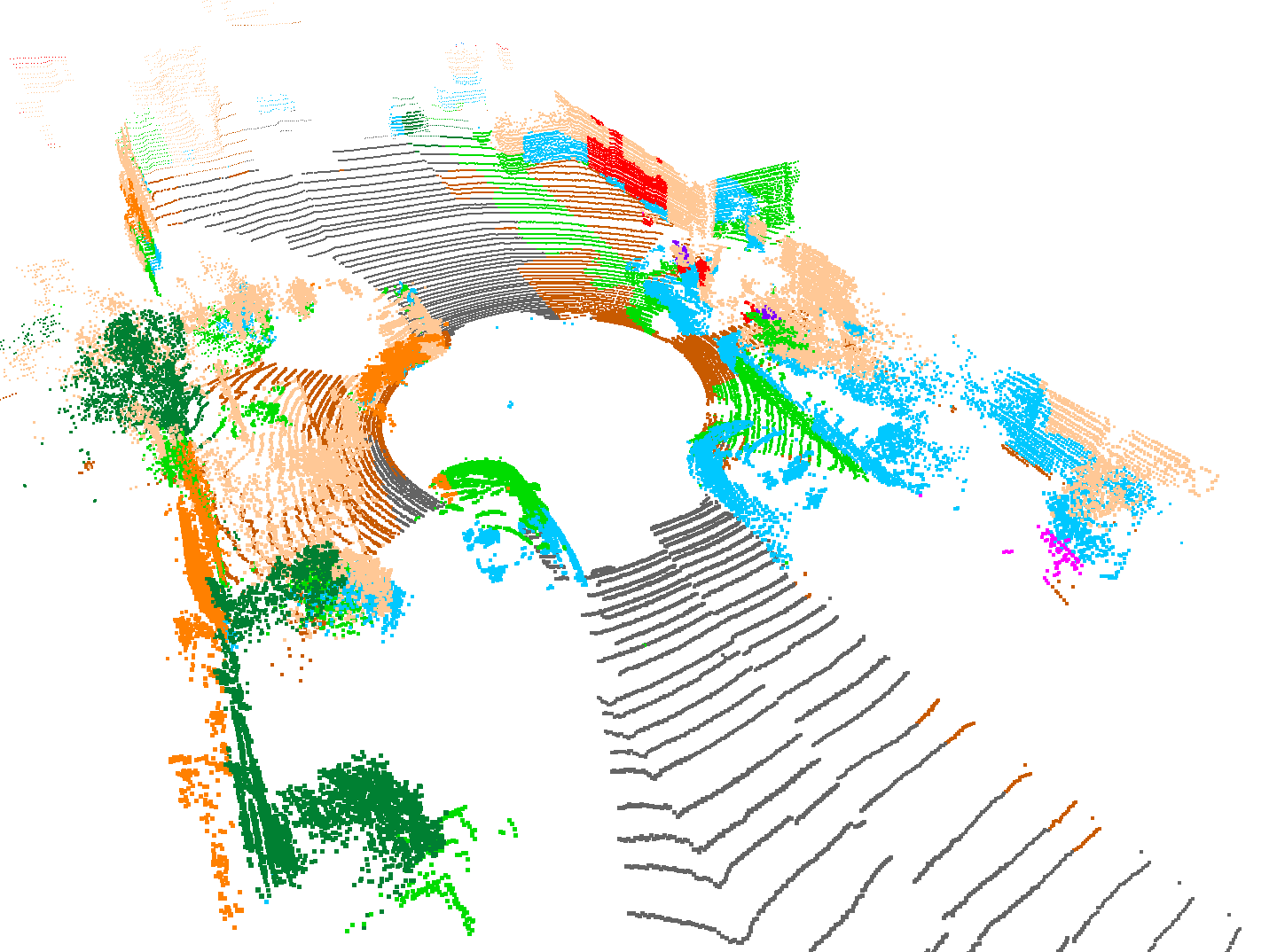}%
\caption{\textbf{Qualitative results on laser point clouds.} Dataset: Velodyne HDL-64E laser scans from KITTI Raw \cite{kitti}.
We trained our model on vKITTI point clouds without color and applied it on real-world laser point clouds. So far, only classes like road, building and car give decent results.}
	\label{fig:quali_results_laser}
\end{figure}

\begin{figure*}[t]
\vspace{-10mm}
\centering
\colorbox{ceiling}{\makebox(35,10){\centering Ceiling}}
\colorbox{floor}{\makebox(26,10){\centering \textcolor{white}{Floor}}}
\colorbox{wall}{\makebox(20,10){\centering Wall}}
\colorbox{beam}{\makebox(25,10){\centering Beam}}
\colorbox{column}{\makebox(35,10){\centering Column}}
\colorbox{window}{\makebox(35,10){\centering Window}}
\colorbox{door}{\makebox(30,10){\centering Door}}
\colorbox{table}{\makebox(25,10){\centering Table}}
\colorbox{chair}{\makebox(25,10){\centering Chair}}
\colorbox{sofa}{\makebox(23,10){\centering Sofa}}
\colorbox{bookcase}{\makebox(45,10){\centering Bookcase}}
\colorbox{board}{\makebox(30,10){\centering Board}}
\colorbox{clutter}{\makebox(35,10){\centering \textcolor{white}{Clutter}}}
\includegraphics[width=0.2\linewidth, trim=100 260 200 100, clip]{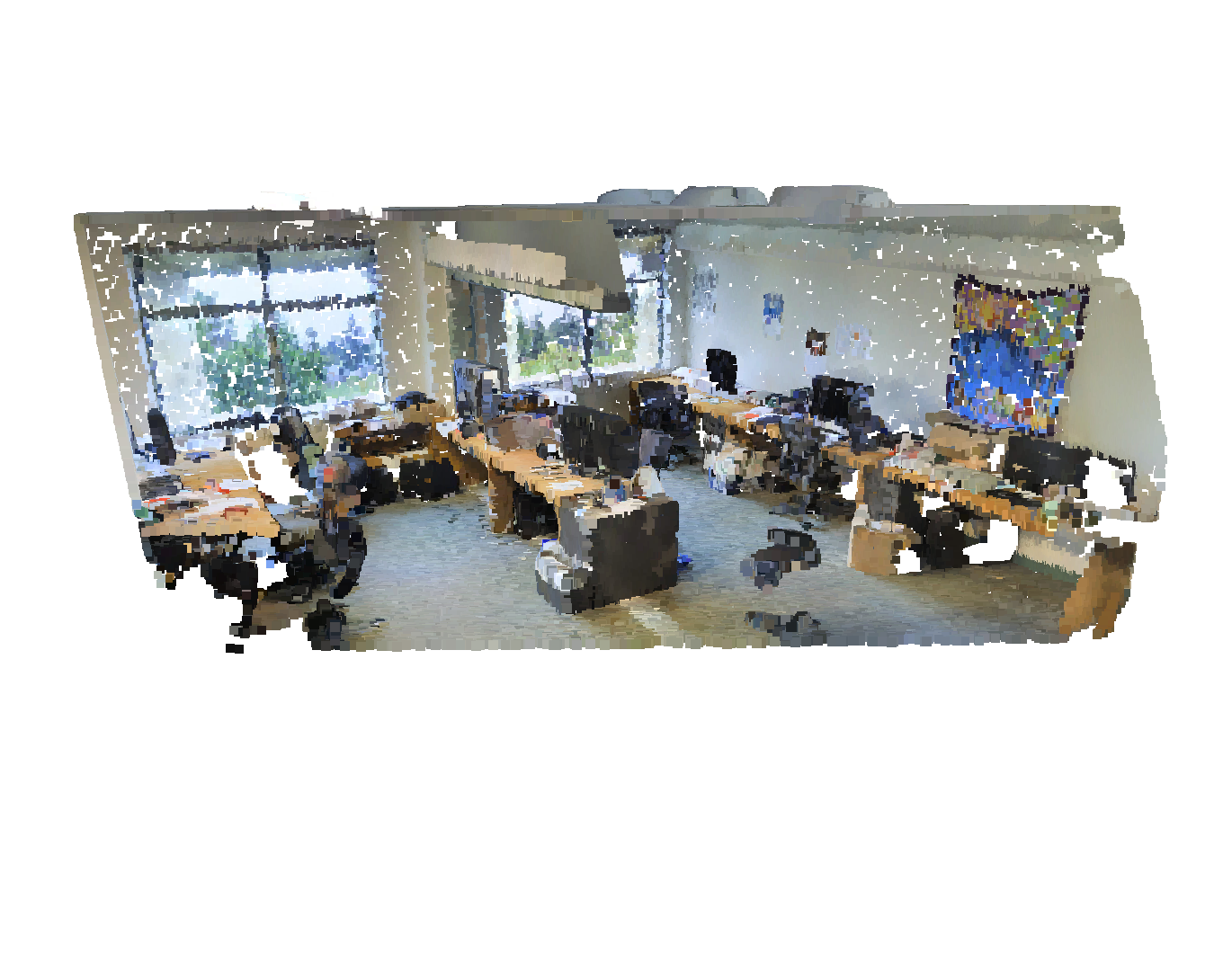}%
\includegraphics[width=0.2\linewidth, trim=100 260 200 100, clip]{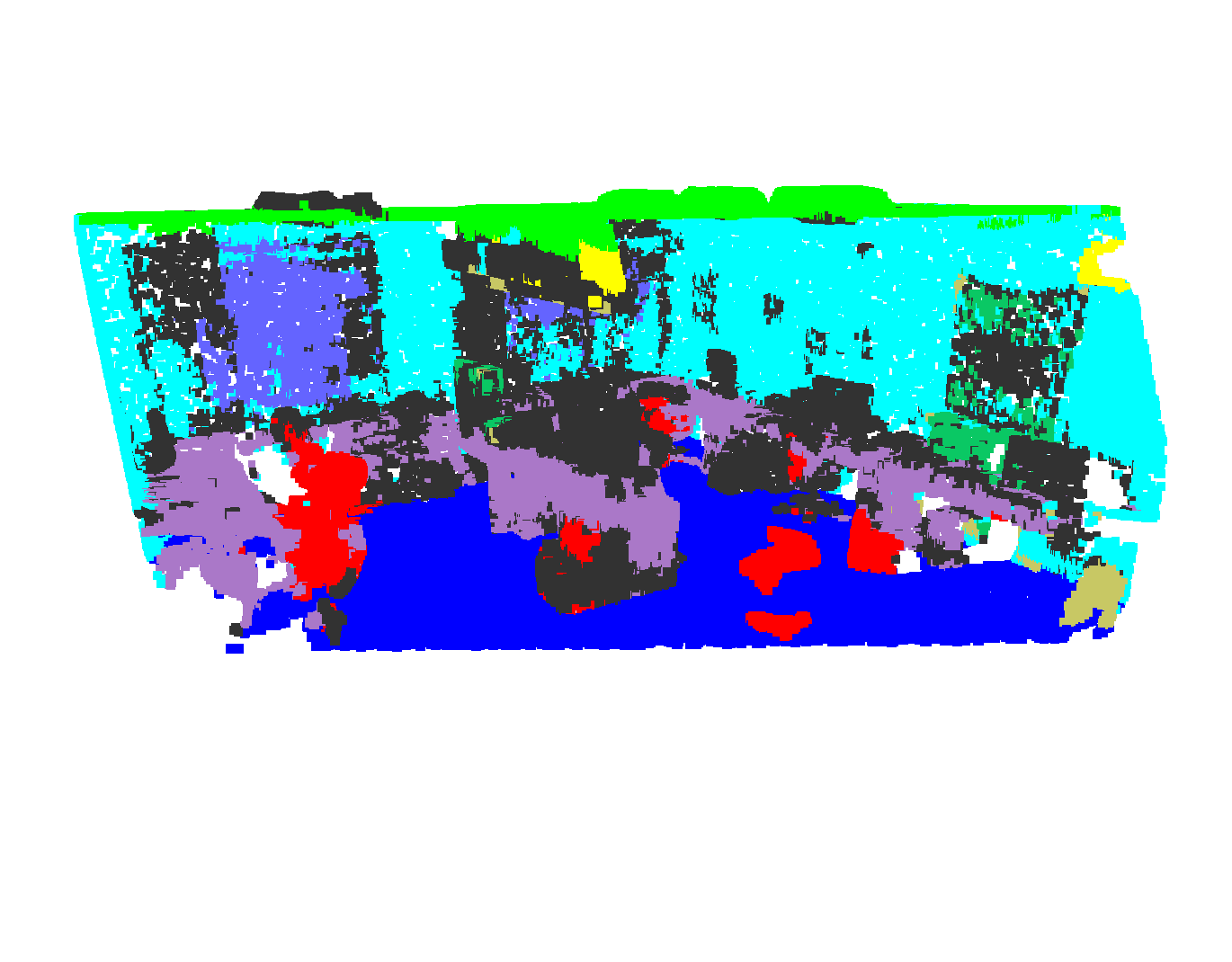}%
\includegraphics[width=0.2\linewidth, trim=100 260 200 100, clip]{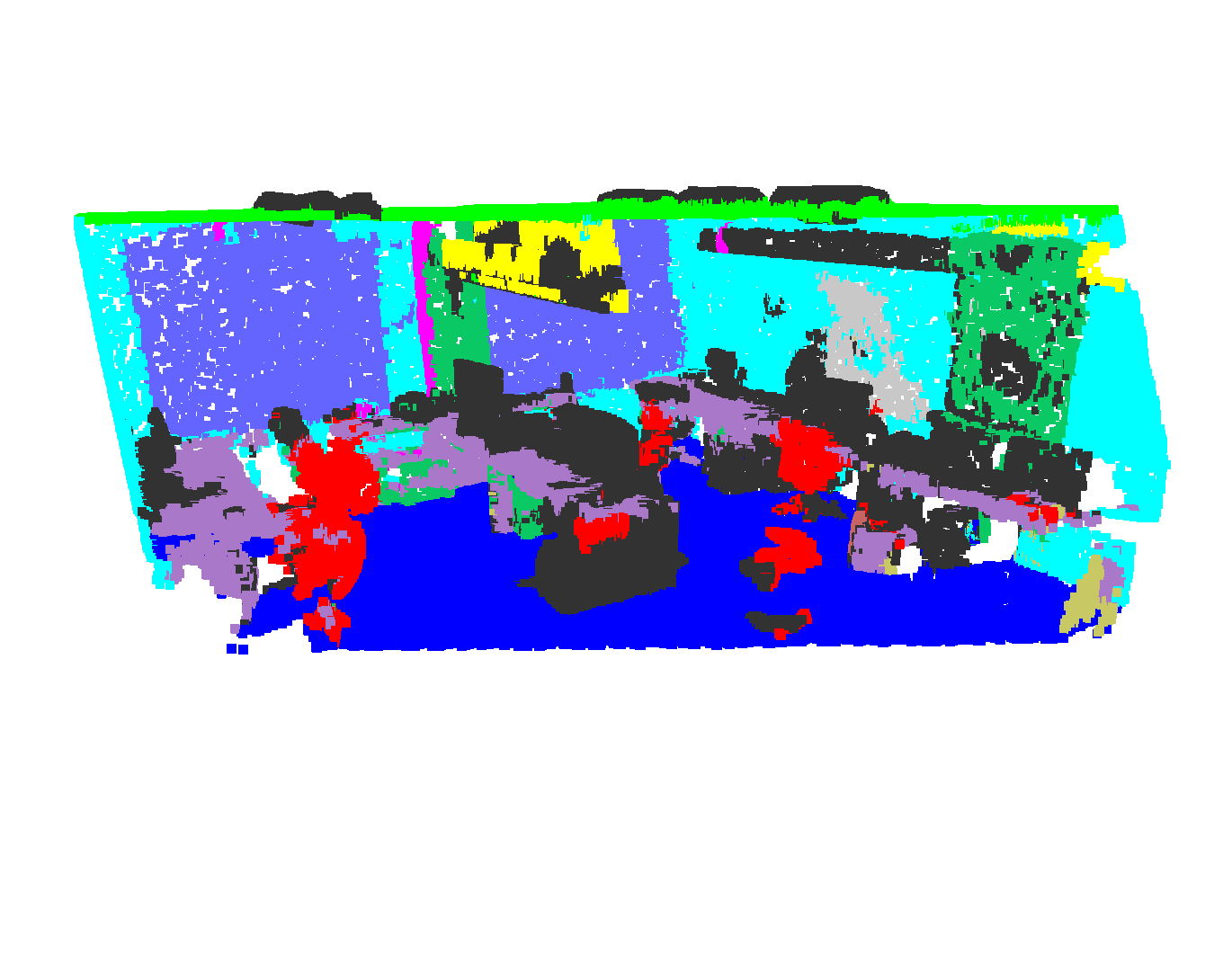}%
\includegraphics[width=0.2\linewidth, trim=100 260 200 100, clip]{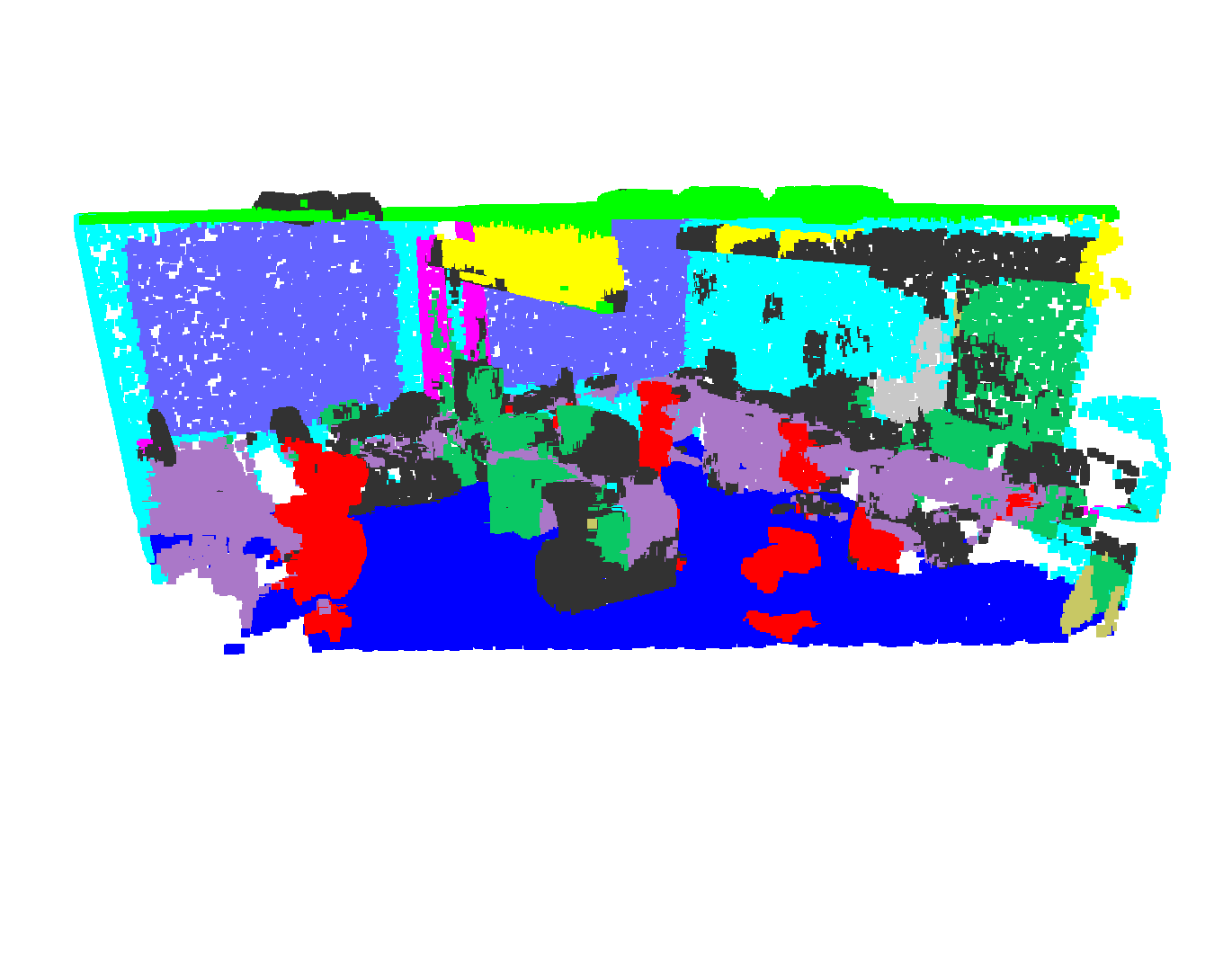}%
\includegraphics[width=0.2\linewidth, trim=100 260 200 100, clip]{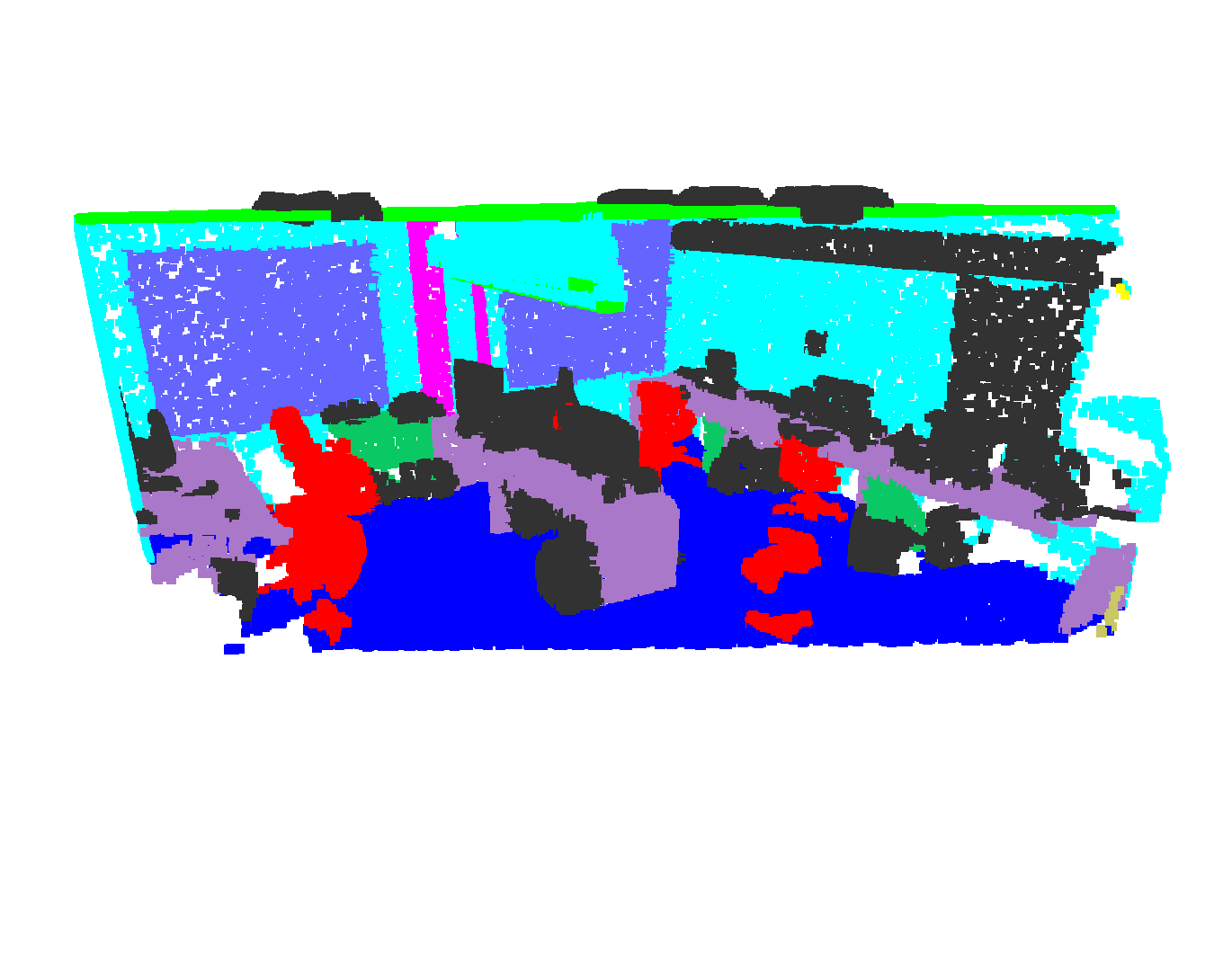}%

\includegraphics[width=0.2\linewidth, trim=100 260 200 0, clip]{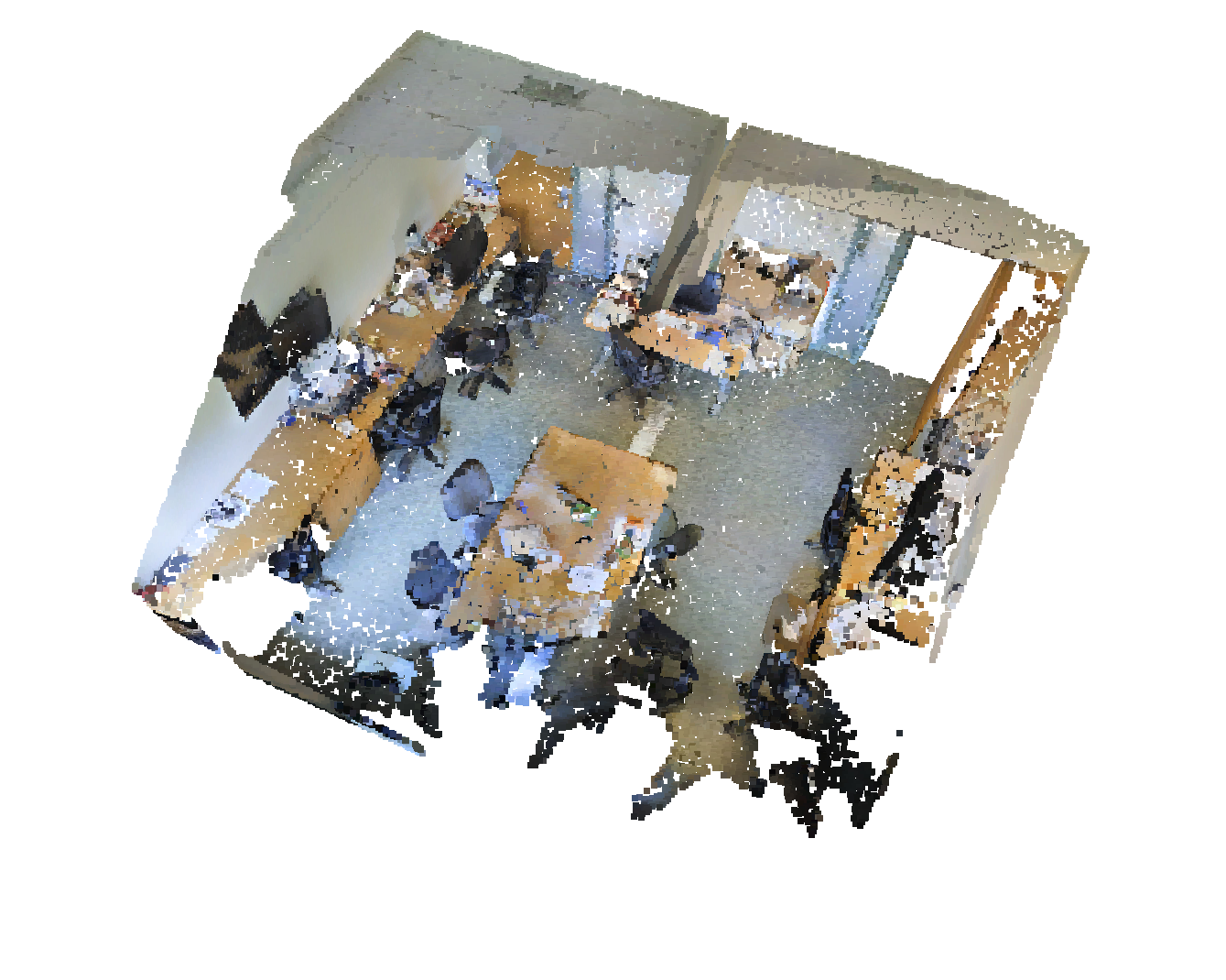}%
\includegraphics[width=0.2\linewidth, trim=100 260 200 0, clip]{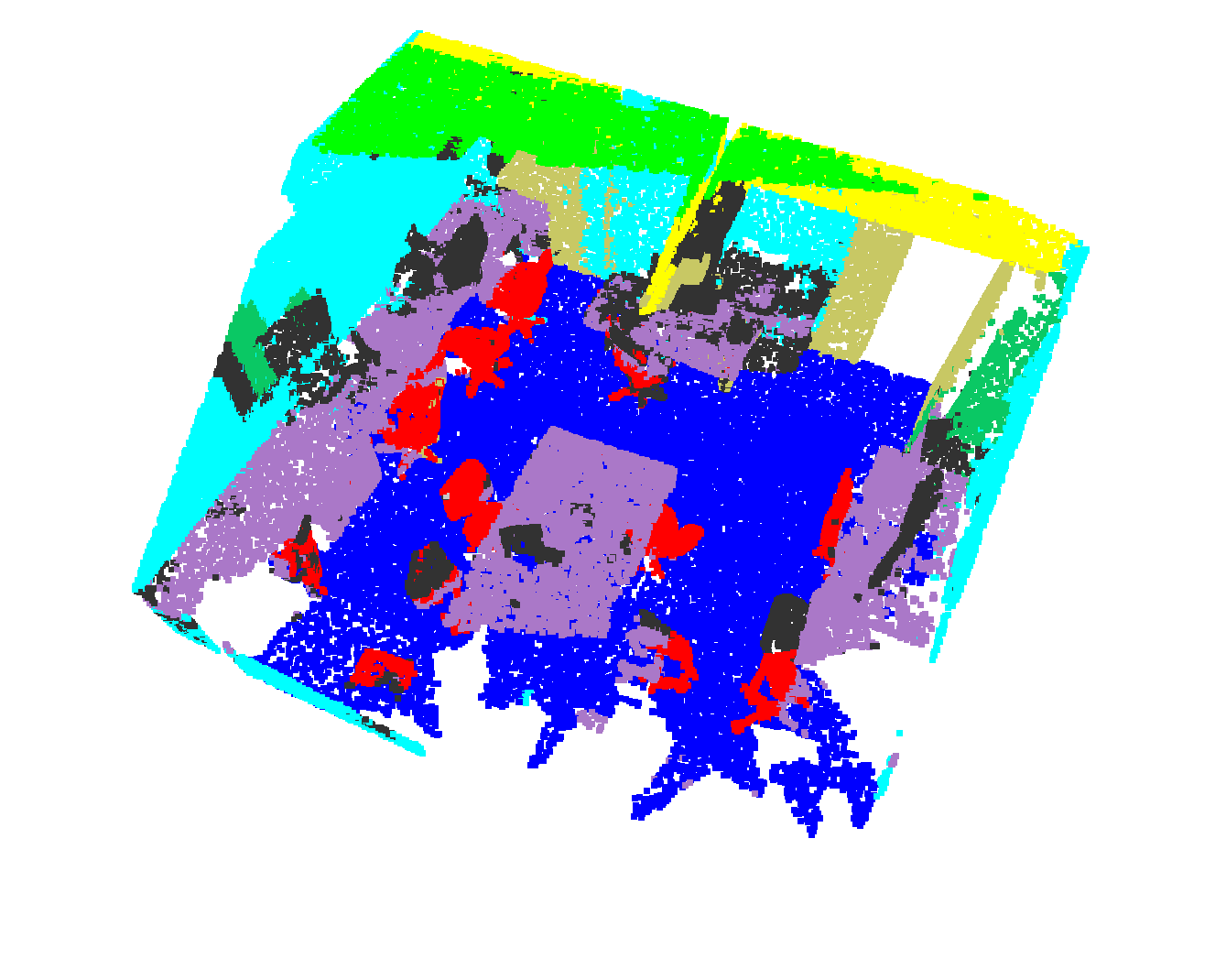}%
\includegraphics[width=0.2\linewidth, trim=100 260 200 0, clip]{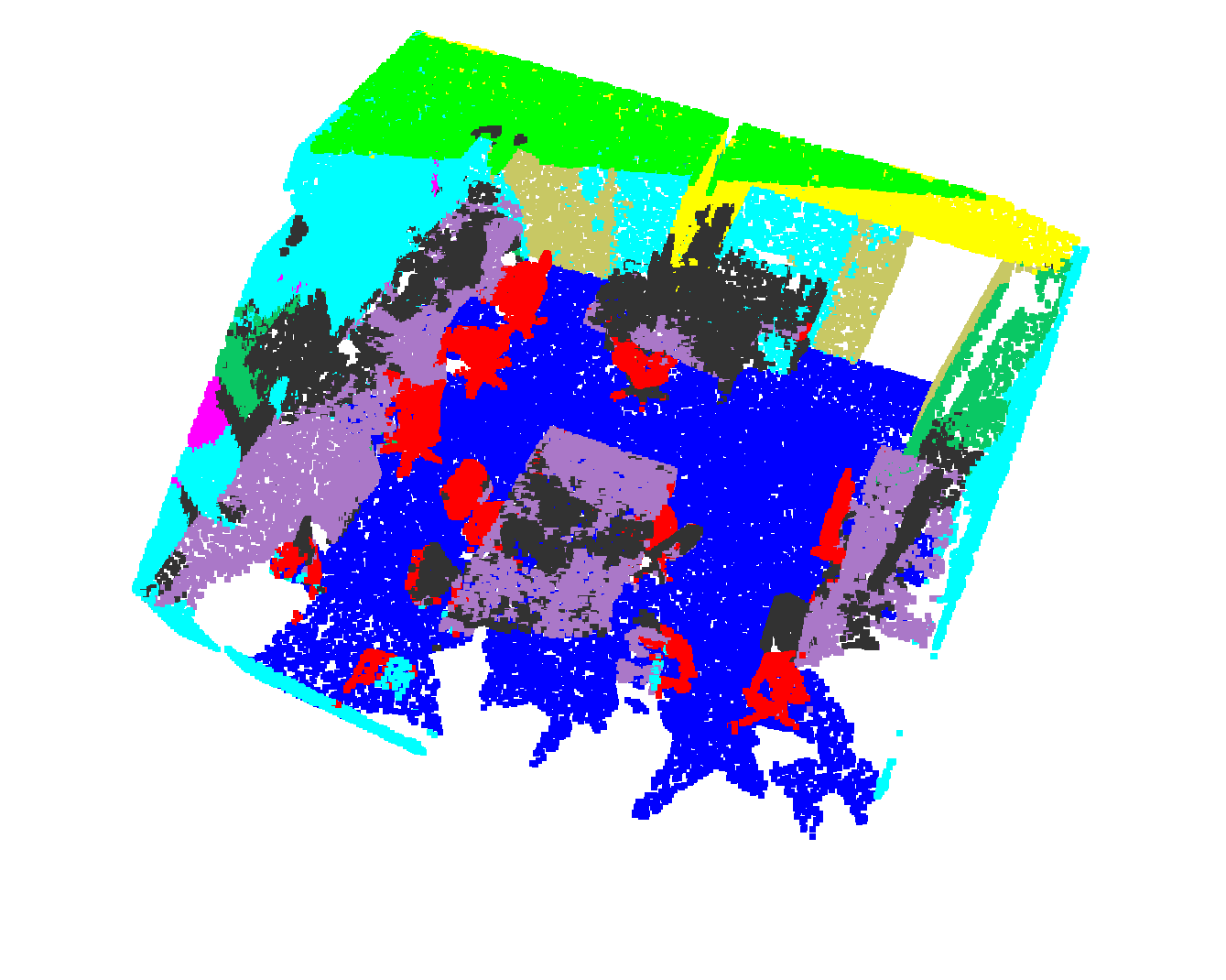}%
\includegraphics[width=0.2\linewidth, trim=100 260 200 0, clip]{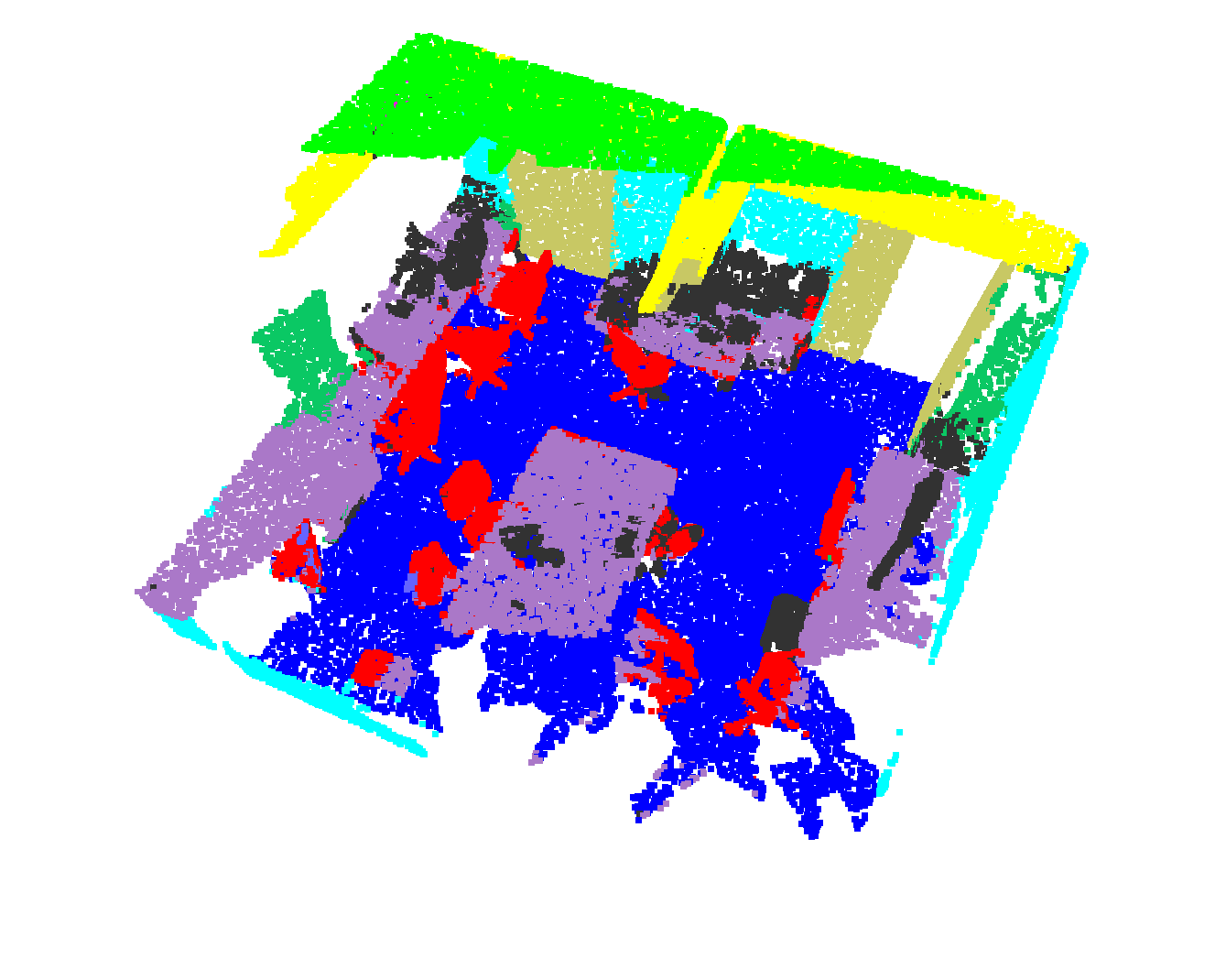}%
\includegraphics[width=0.2\linewidth, trim=100 260 200 0, clip]{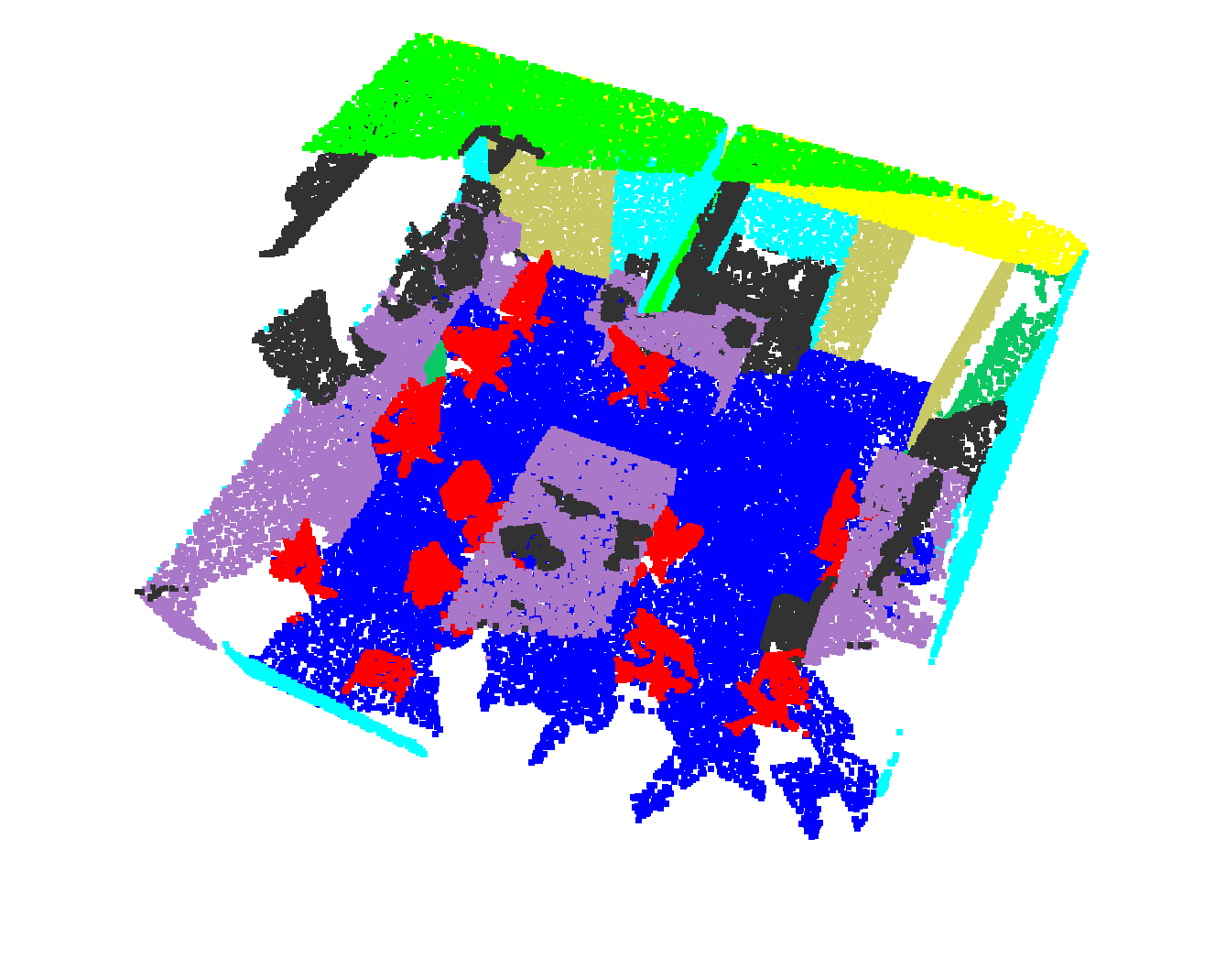}%

\includegraphics[width=0.2\linewidth, trim=100 260 200 60, clip]{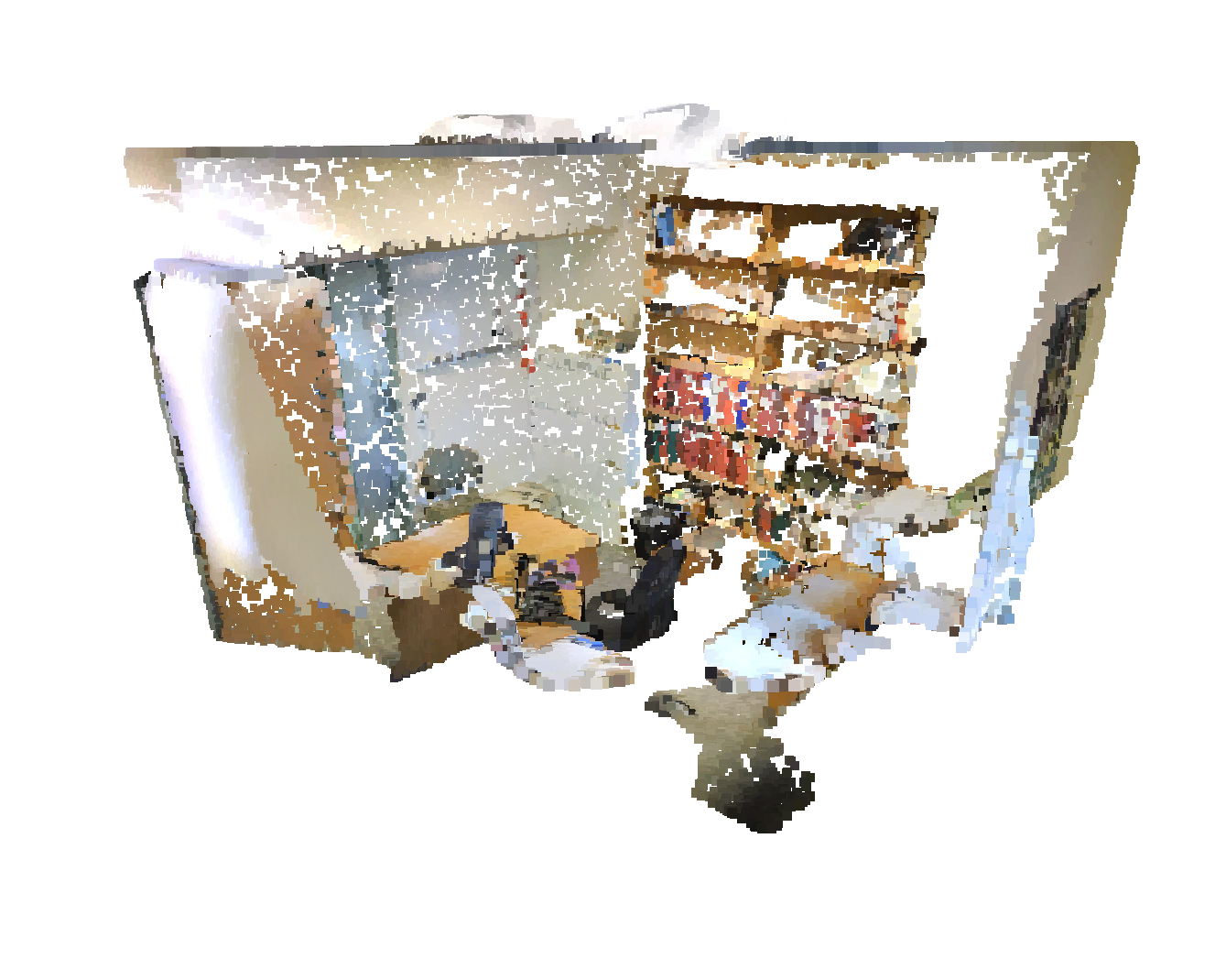}%
\includegraphics[width=0.2\linewidth, trim=100 260 200 60, clip]{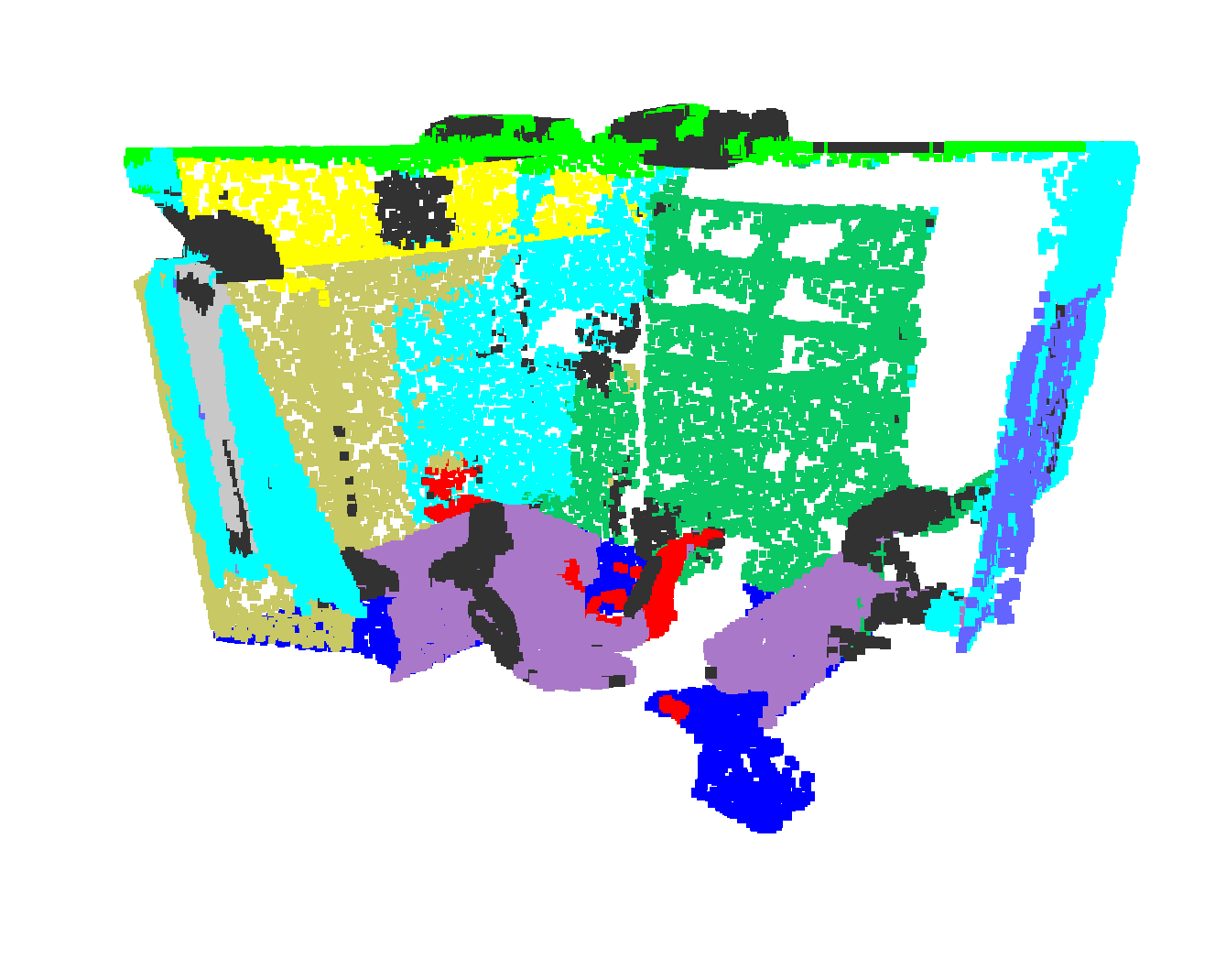}%
\includegraphics[width=0.2\linewidth, trim=100 260 200 60, clip]{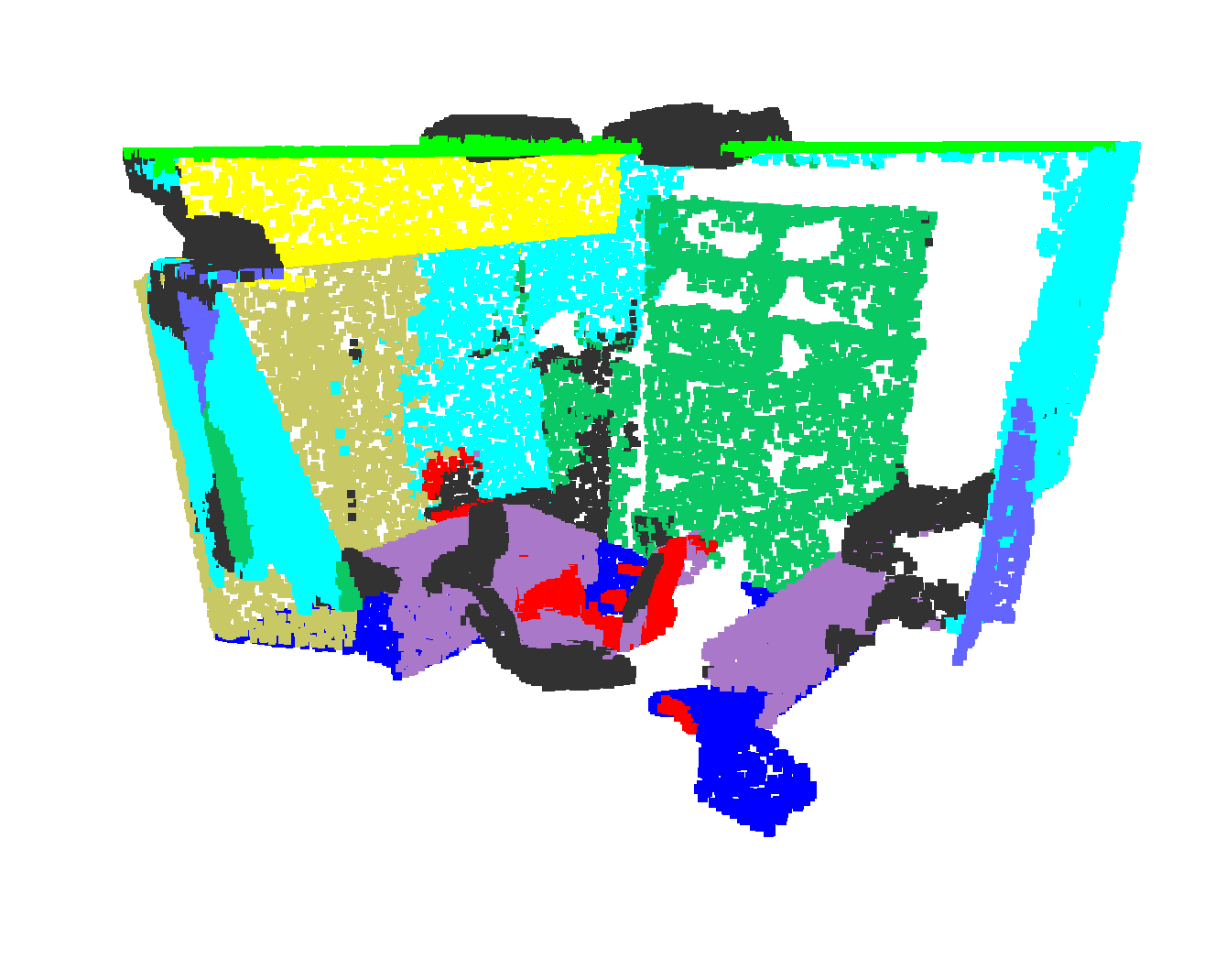}%
\includegraphics[width=0.2\linewidth, trim=100 260 200 60, clip]{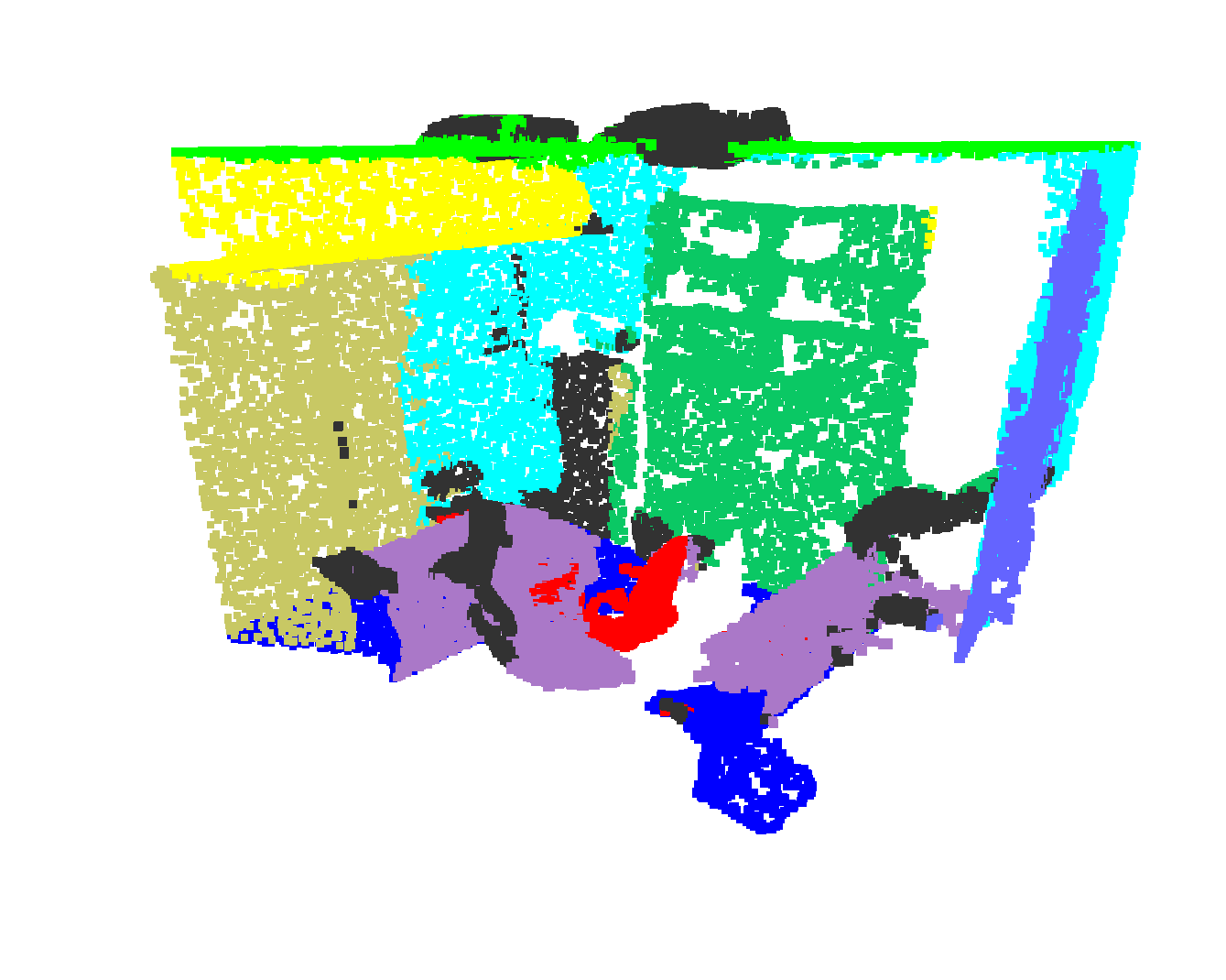}%
\includegraphics[width=0.2\linewidth, trim=100 260 200 60, clip]{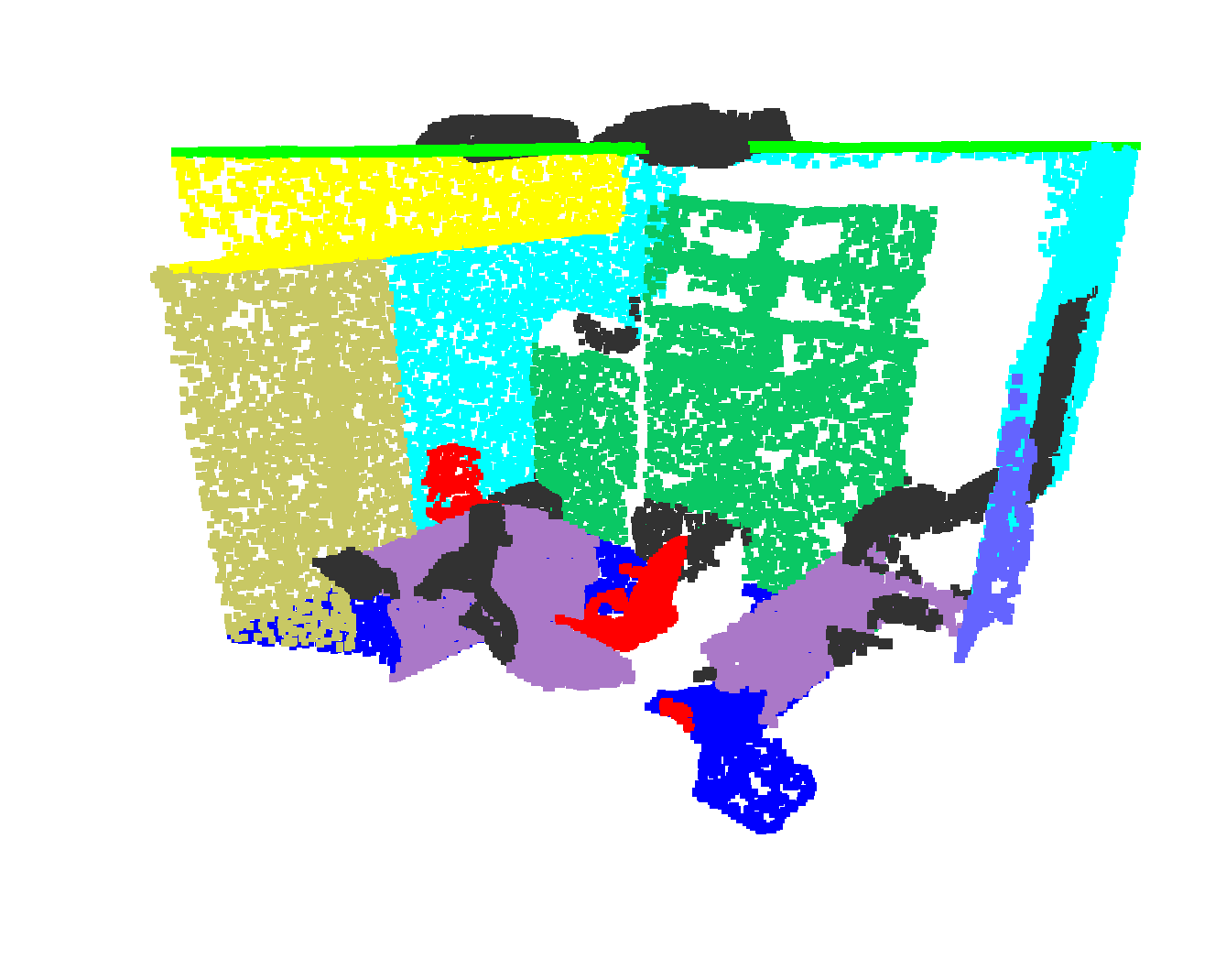}%

\includegraphics[width=0.2\linewidth, trim=100 260 200 150, clip]{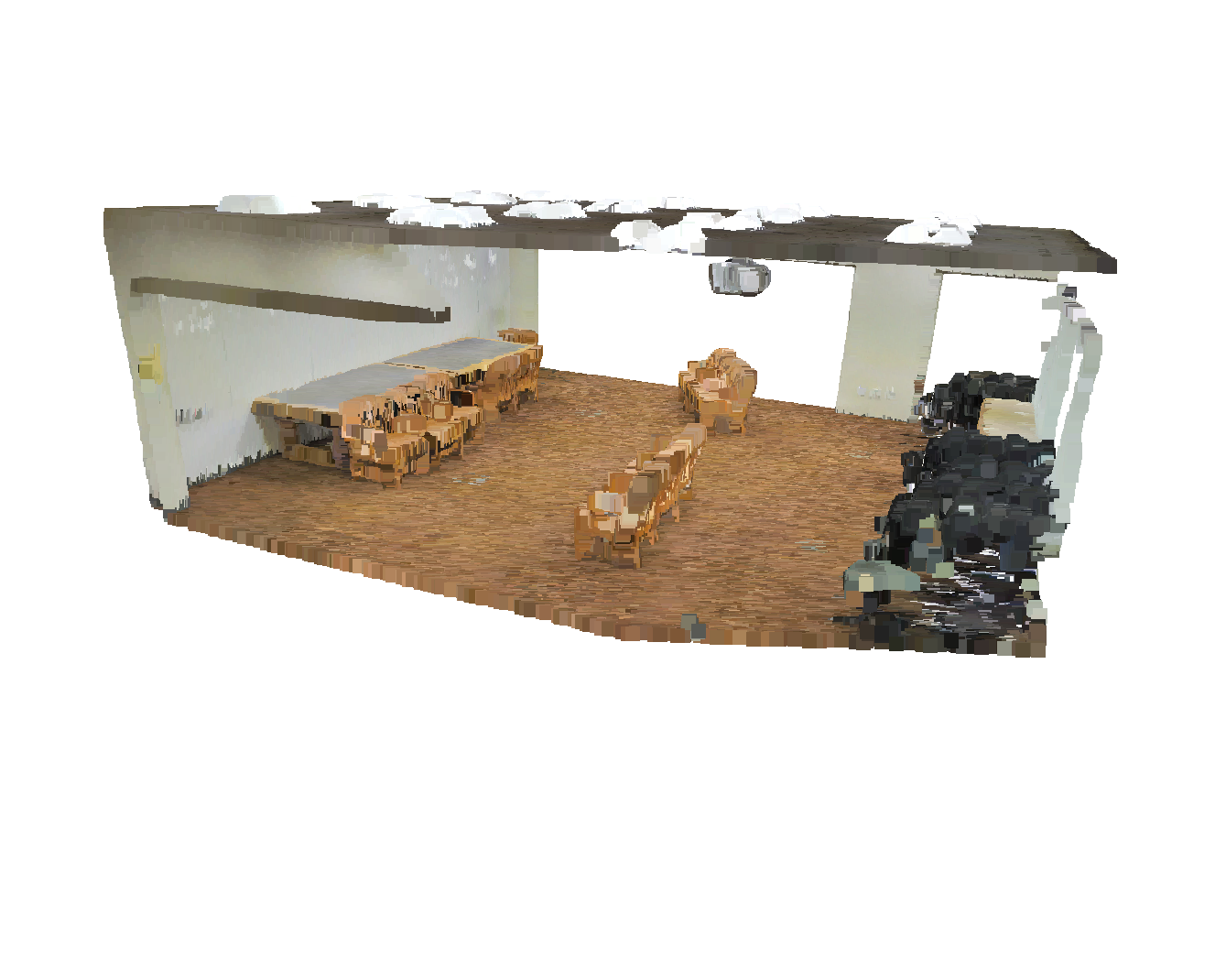}%
\includegraphics[width=0.2\linewidth, trim=100 260 200 150, clip]{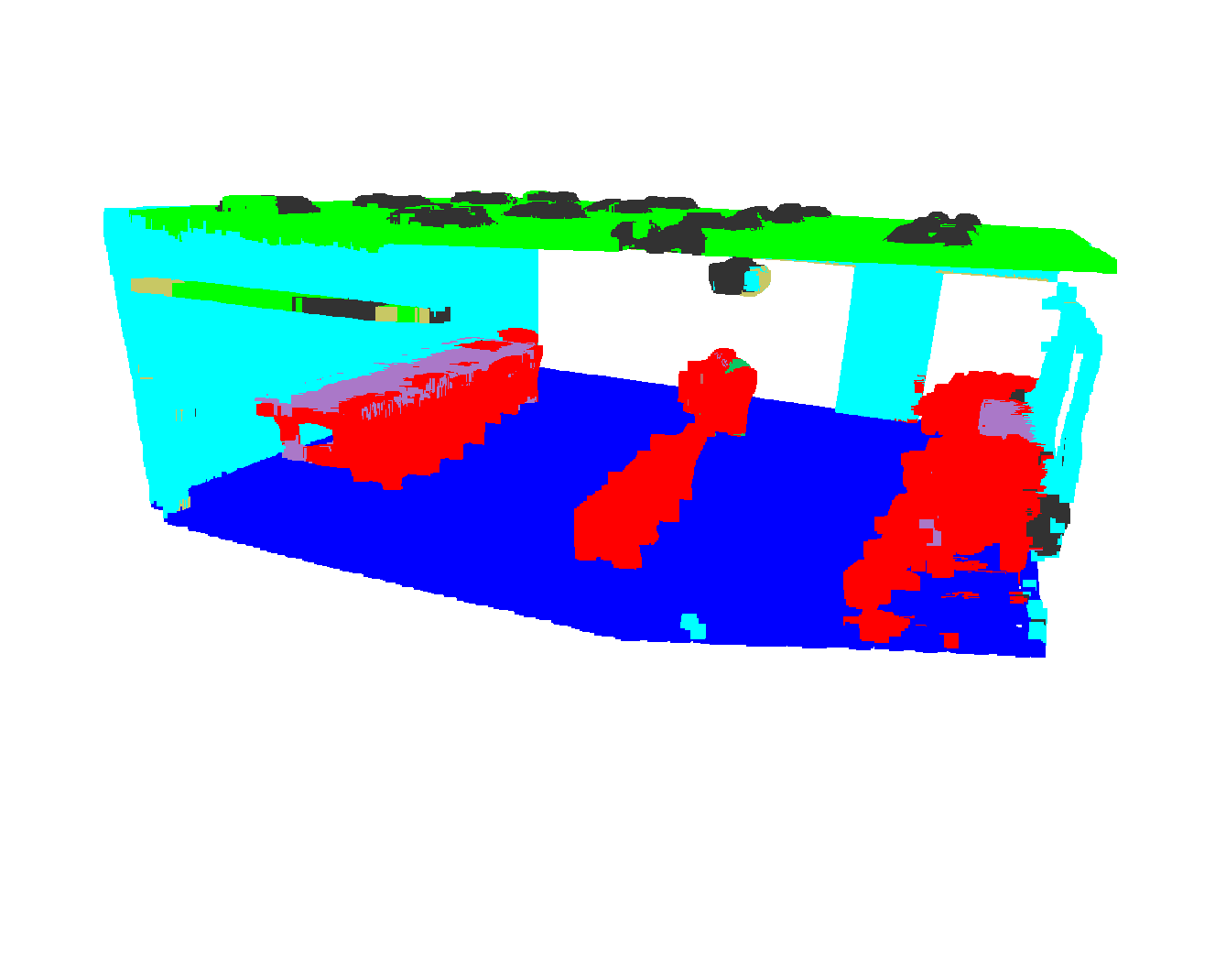}%
\includegraphics[width=0.2\linewidth, trim=100 260 200 150, clip]{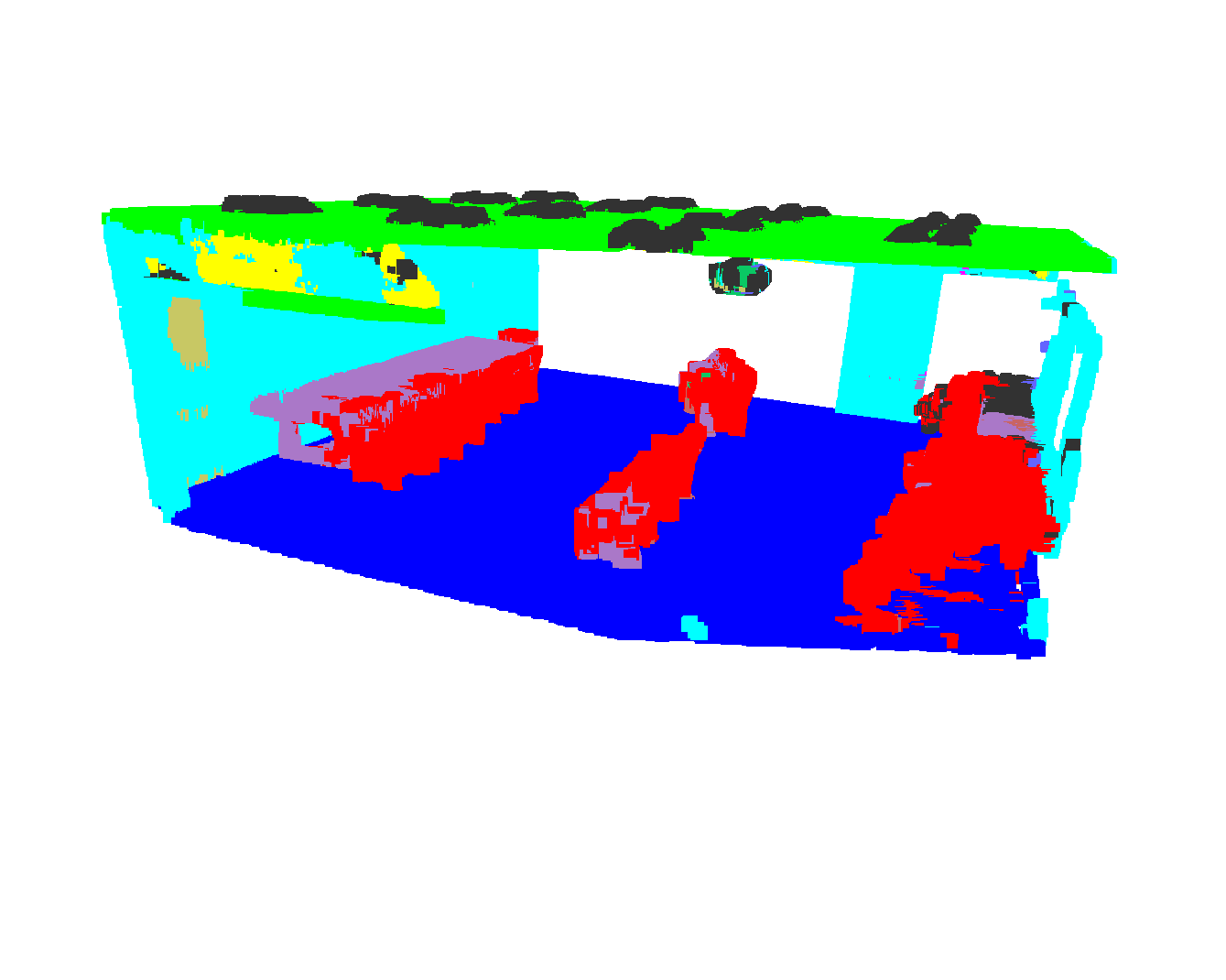}%
\includegraphics[width=0.2\linewidth, trim=100 260 200 150, clip]{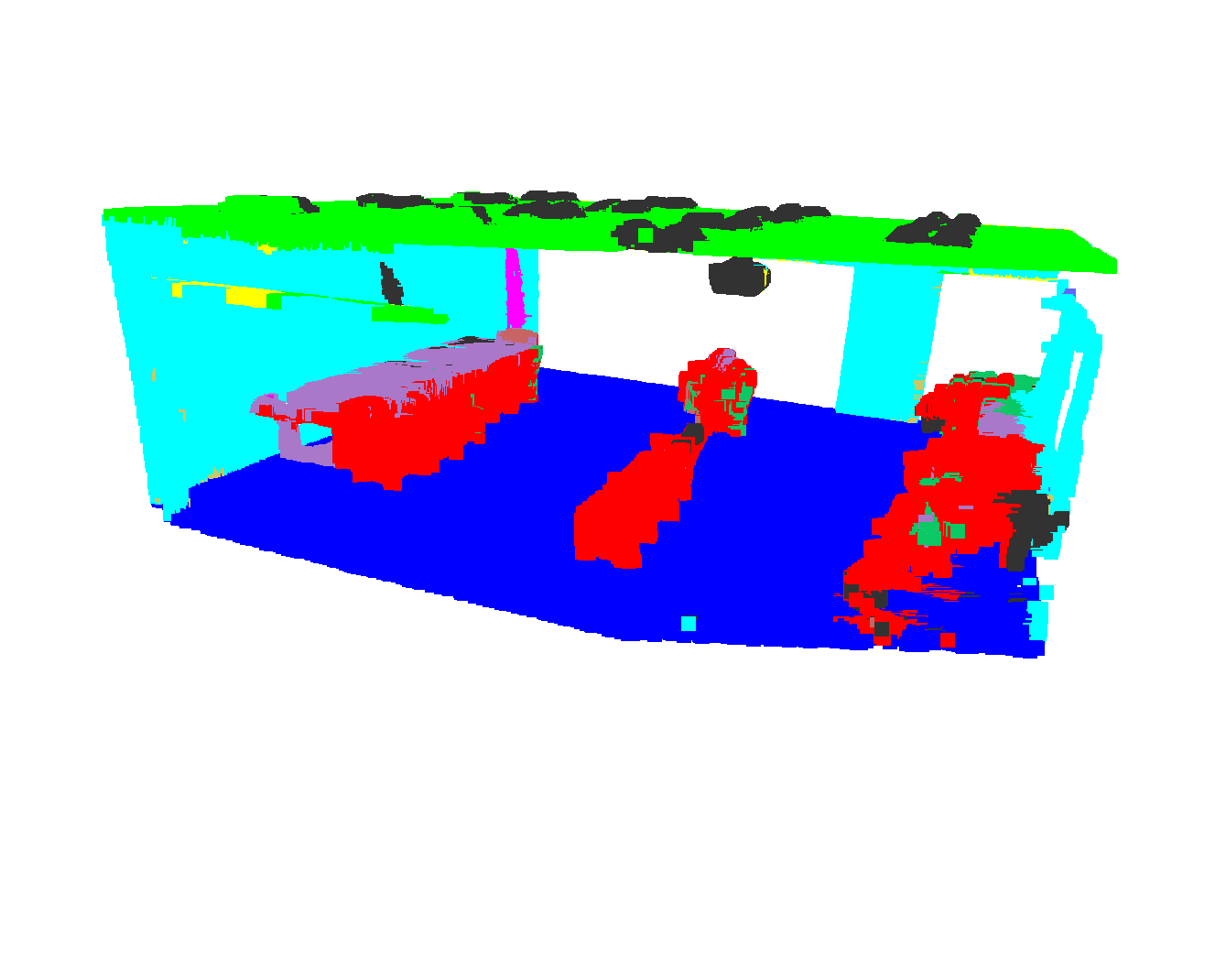}%
\includegraphics[width=0.2\linewidth, trim=100 260 200 150, clip]{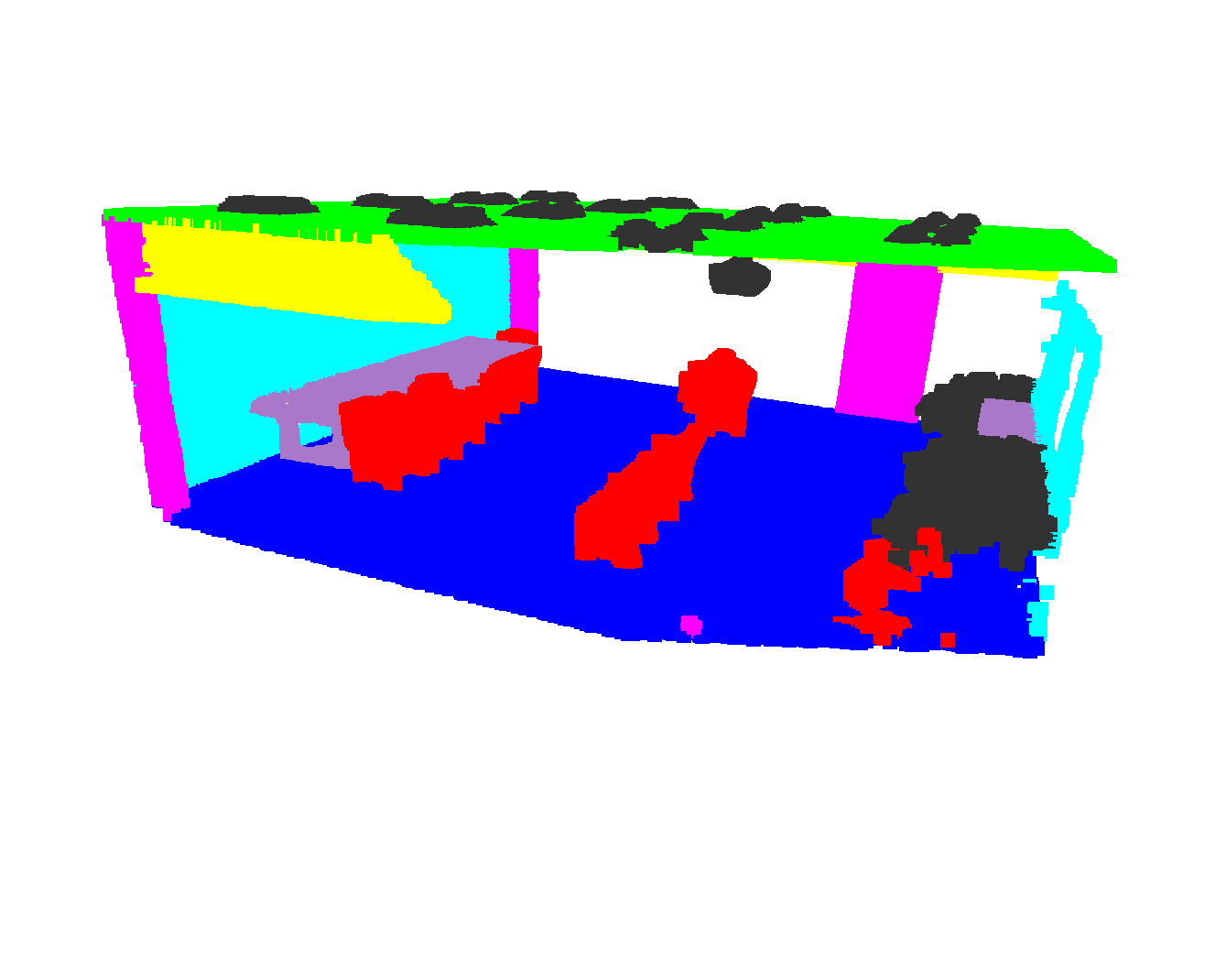}%

\includegraphics[width=0.2\linewidth, trim=0 0 0 100, clip]{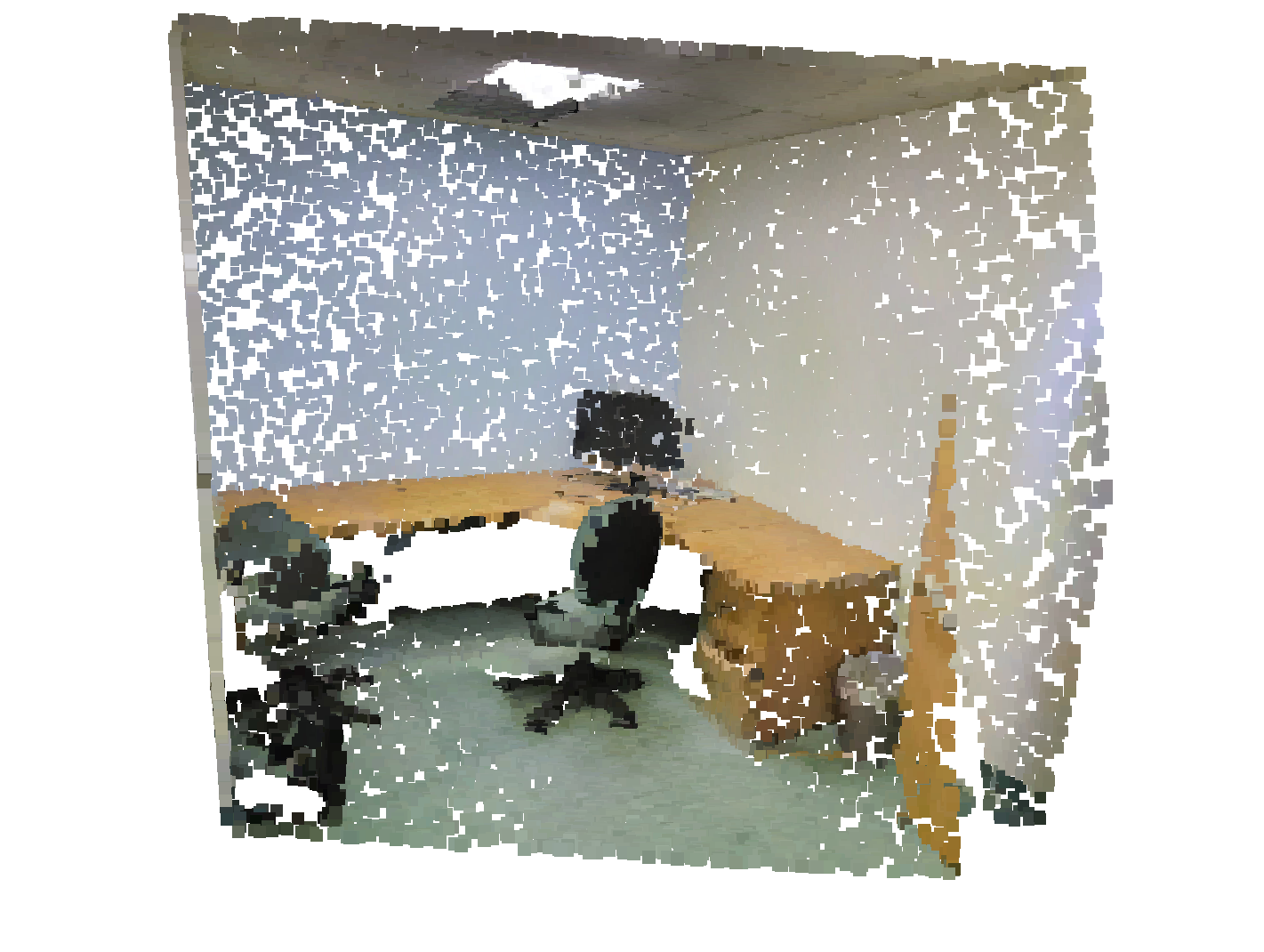}%
\includegraphics[width=0.2\linewidth, trim=0 0 0 100, clip]{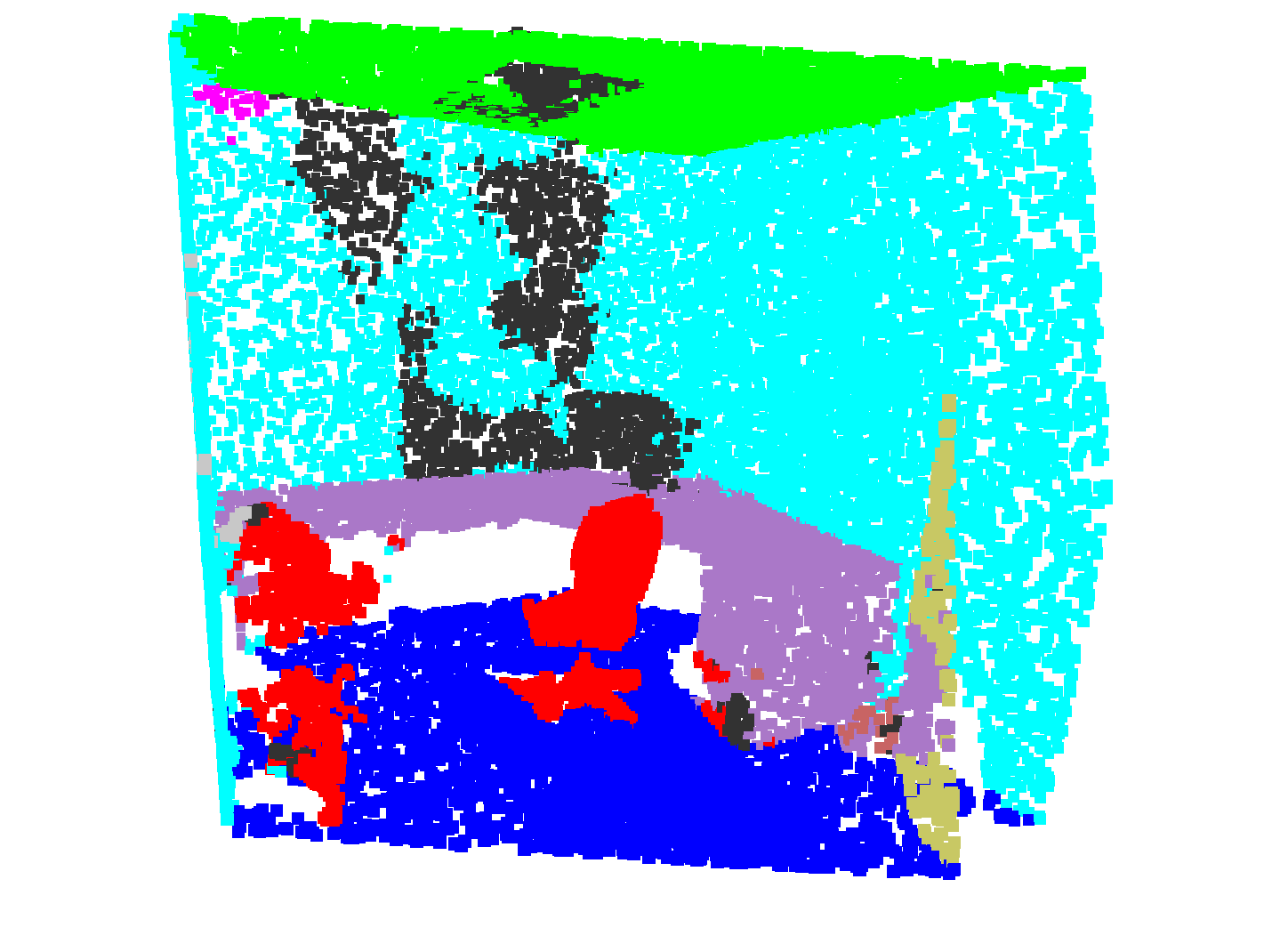}%
\includegraphics[width=0.2\linewidth, trim=0 0 0 100, clip]{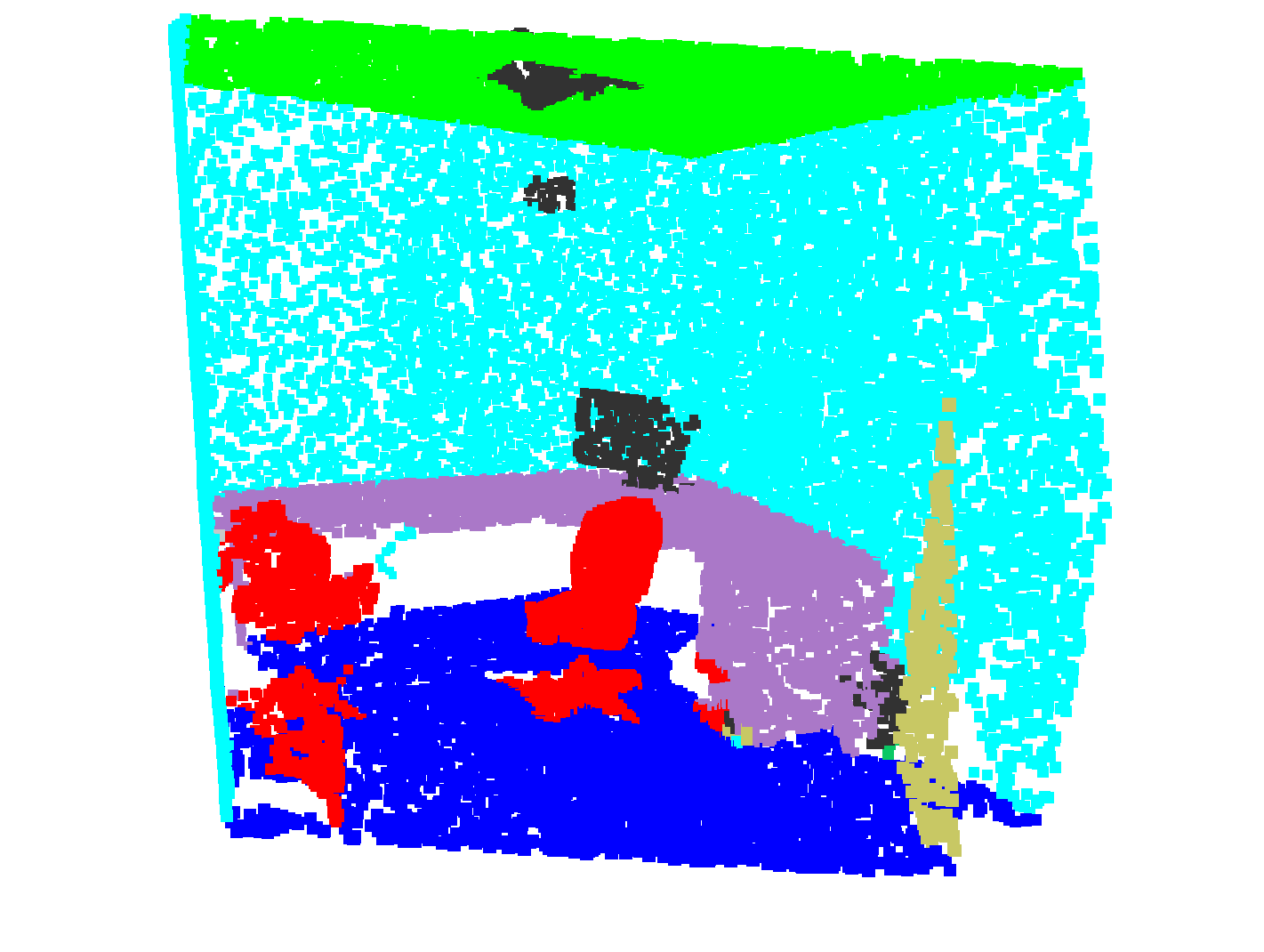}%
\includegraphics[width=0.2\linewidth, trim=0 0 0 100, clip]{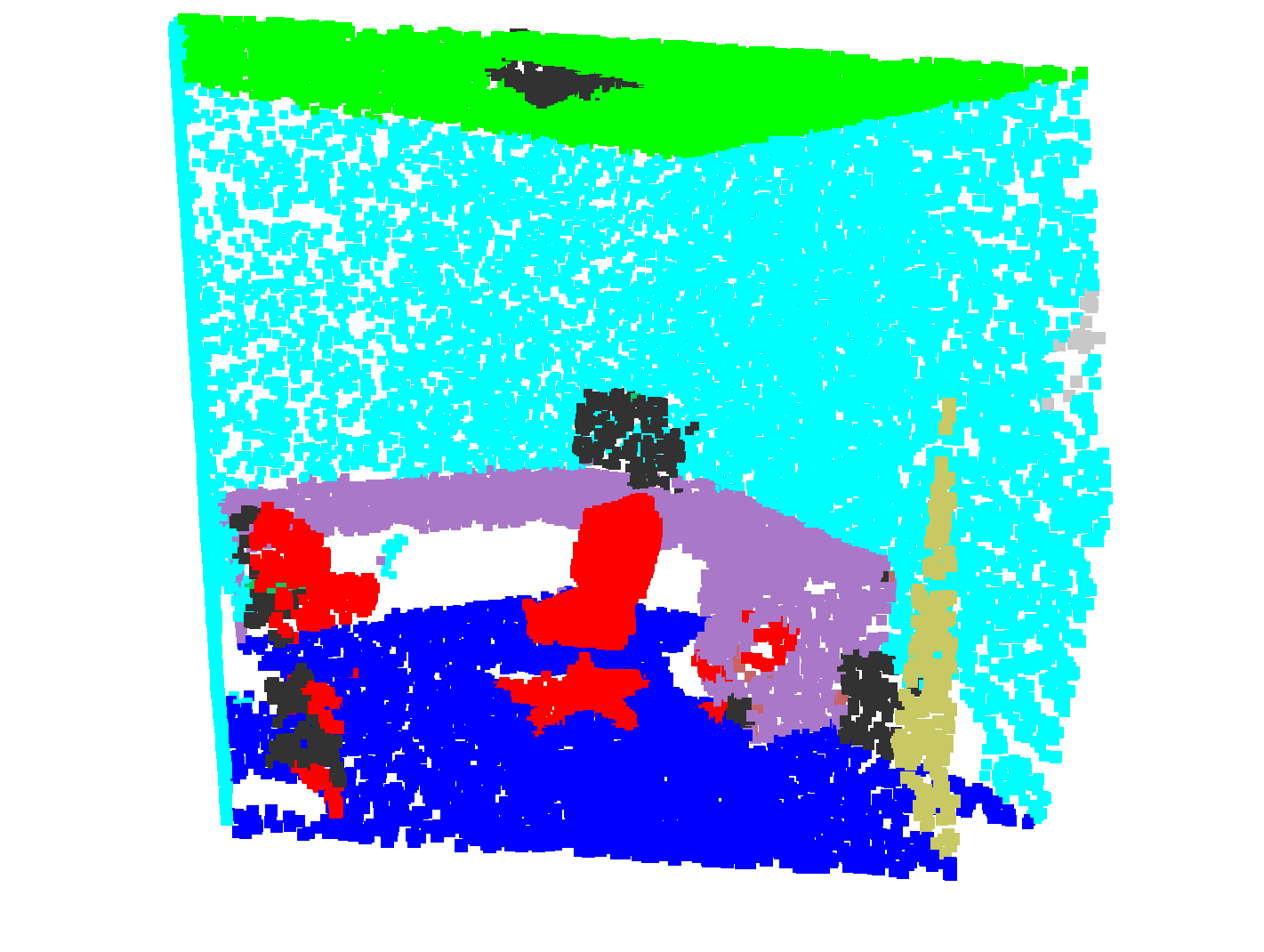}%
\includegraphics[width=0.2\linewidth, trim=0 0 0 100, clip]{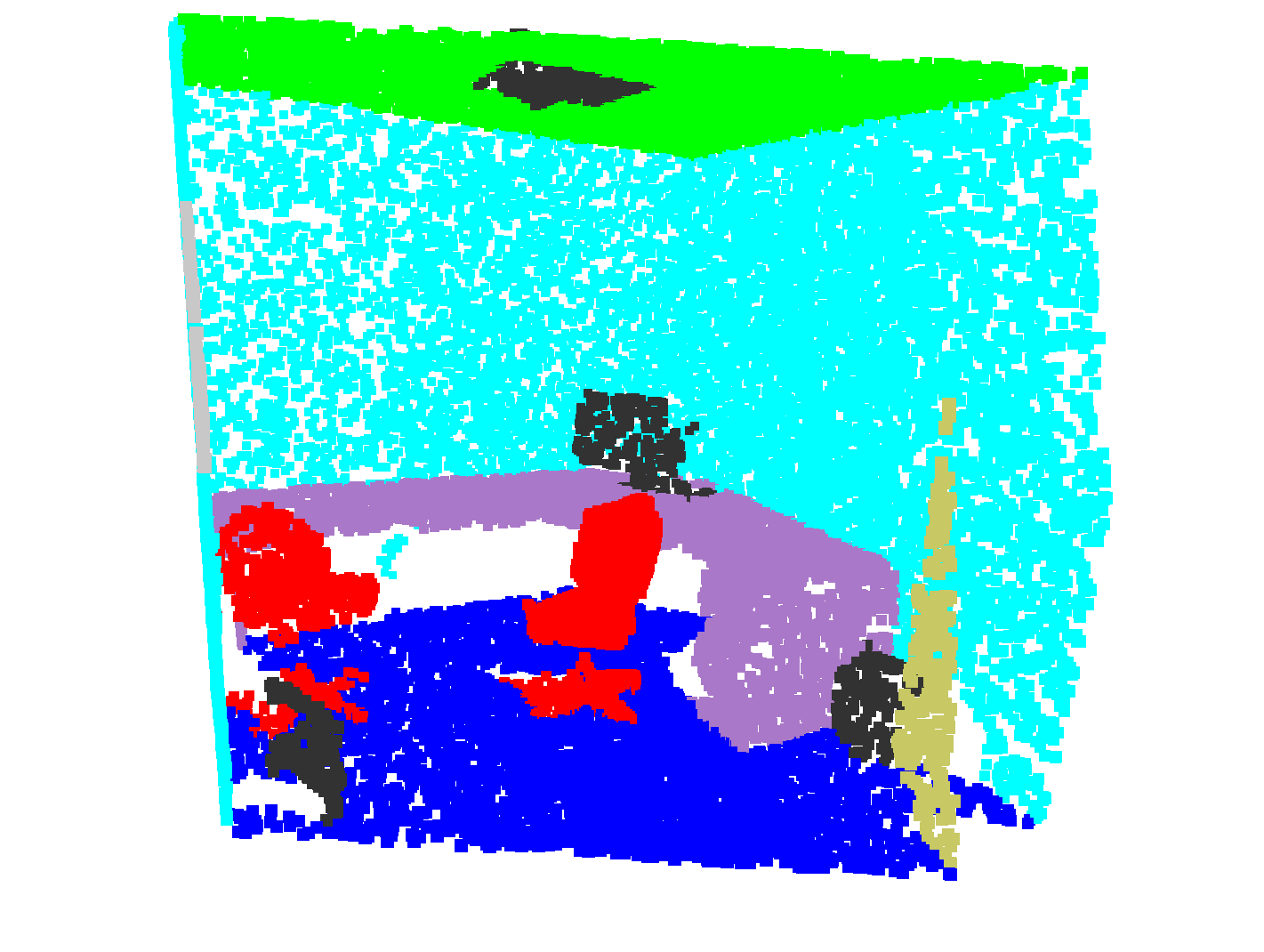}%

\includegraphics[width=0.15\linewidth, trim=200 0 350 200, clip]{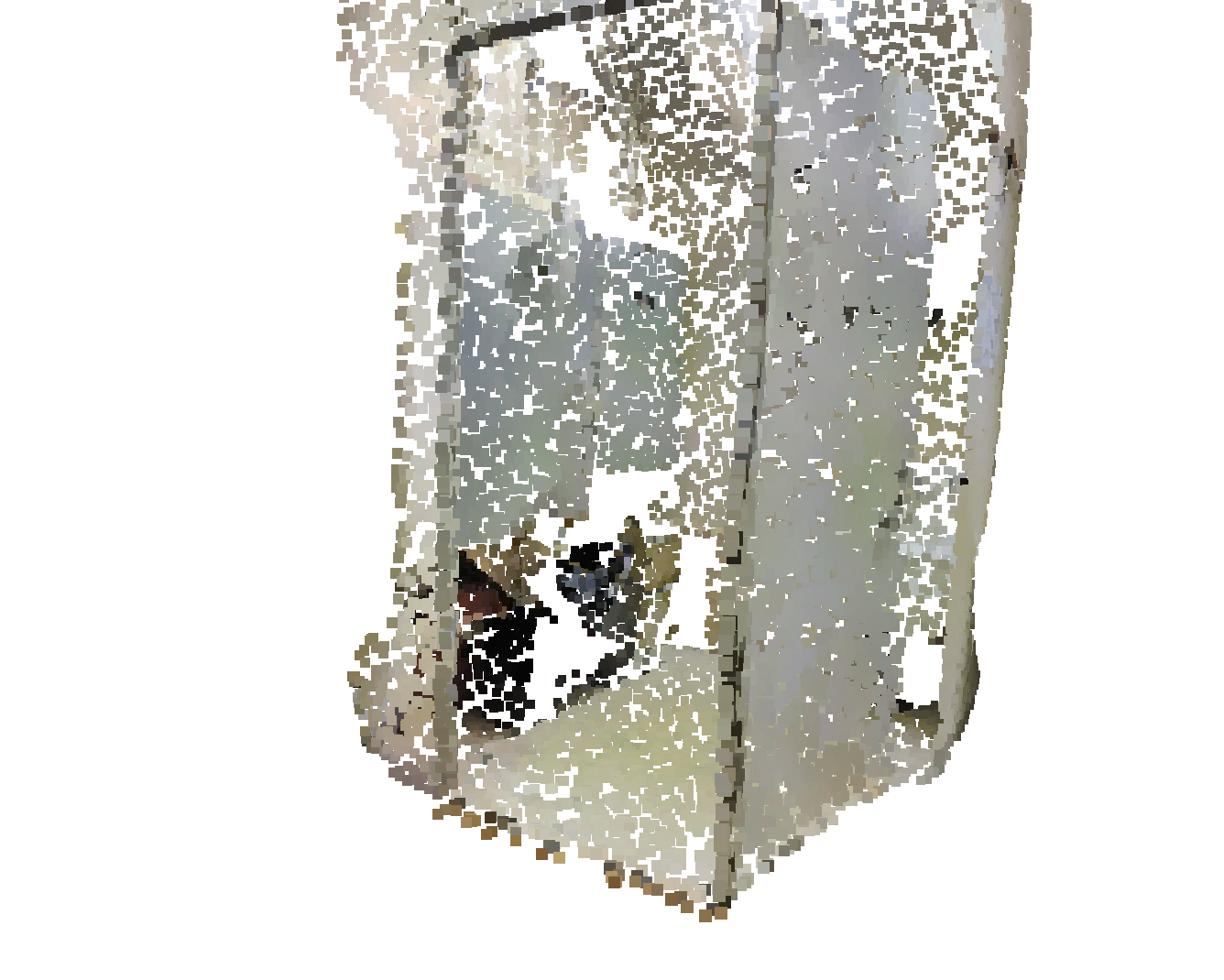} \hspace{8mm}%
\includegraphics[width=0.15\linewidth, trim=200 0 350 200, clip]{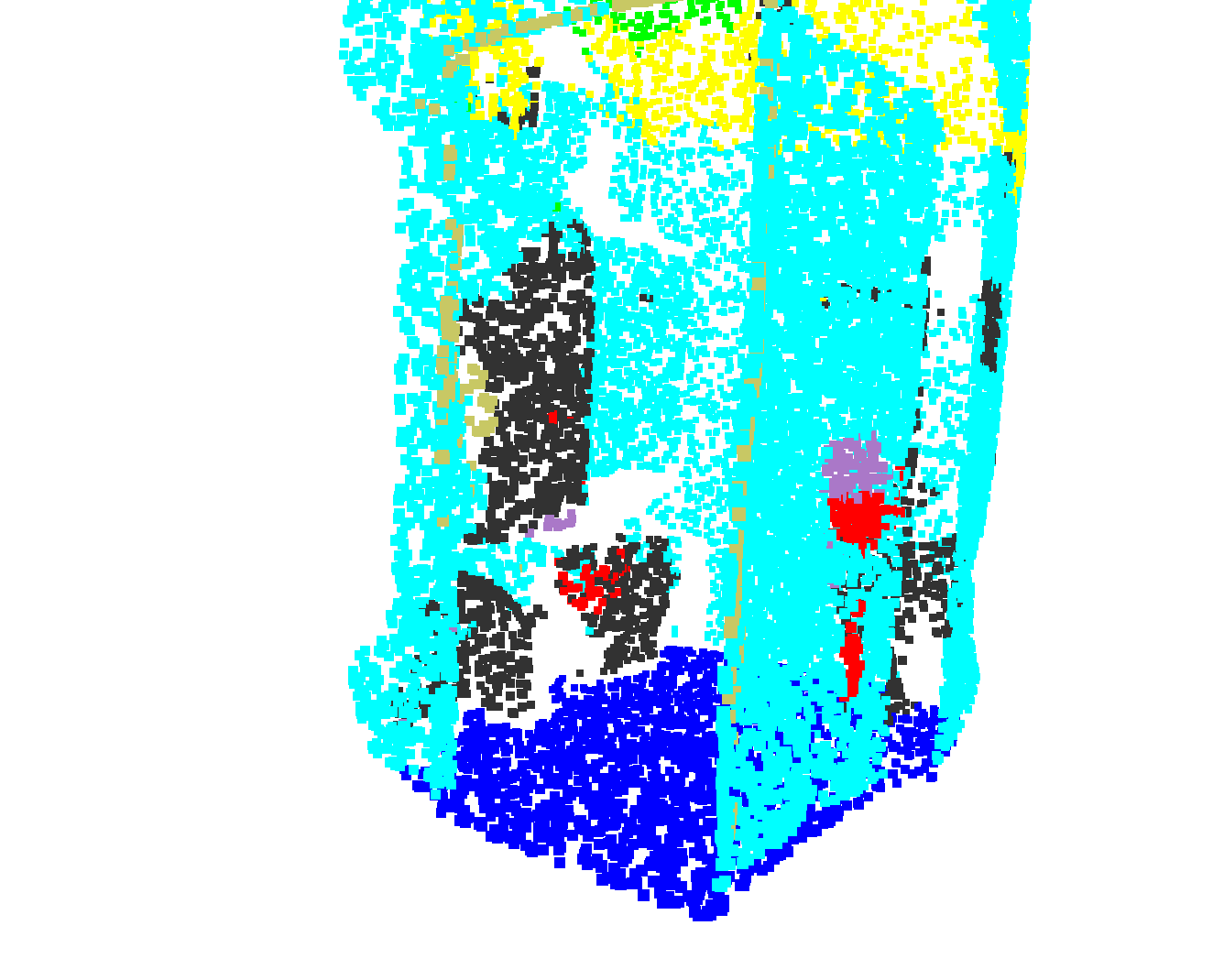} \hspace{8mm}%
\includegraphics[width=0.15\linewidth, trim=200 0 350 200, clip]{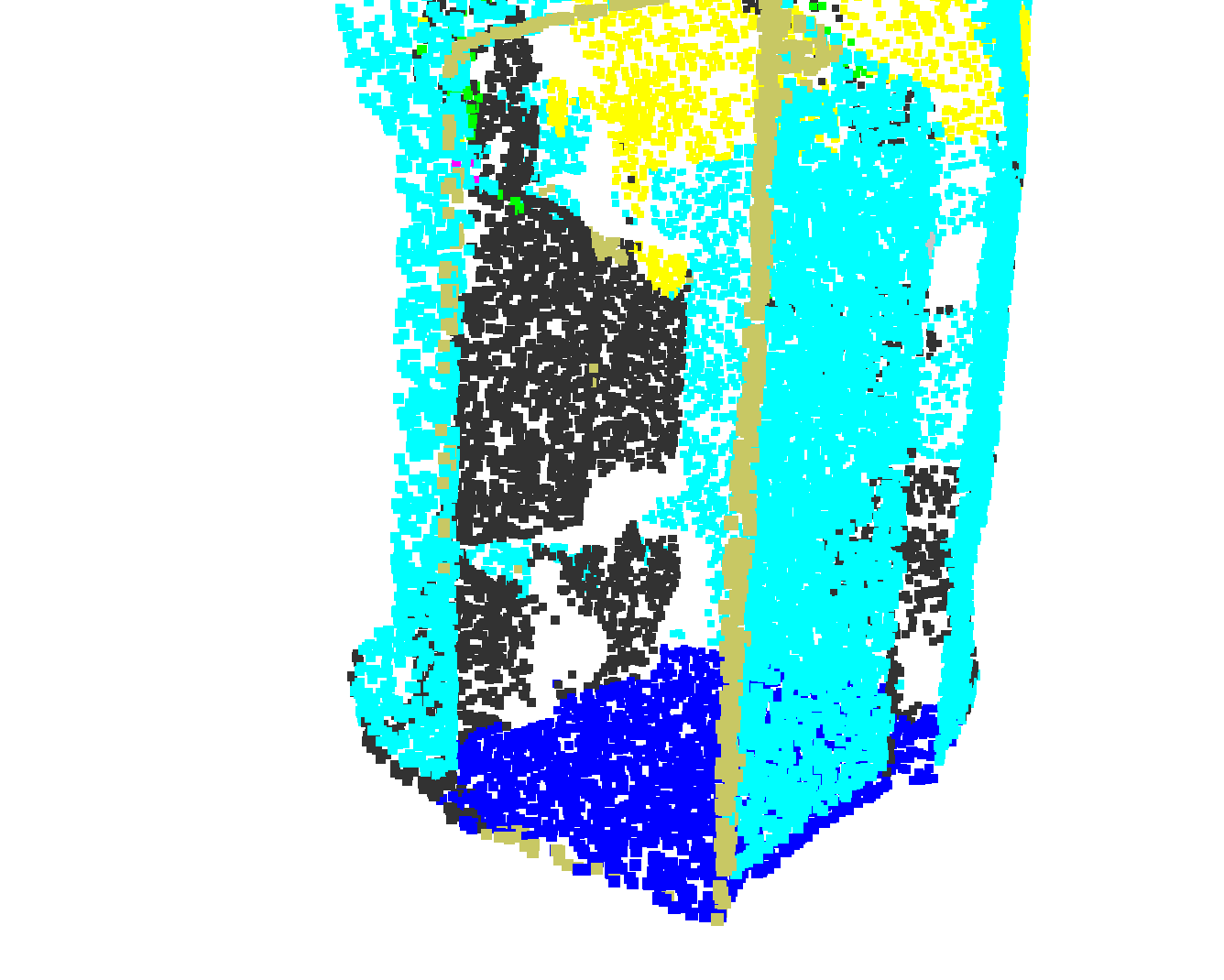} \hspace{8mm}%
\includegraphics[width=0.15\linewidth, trim=200 0 350 200, clip]{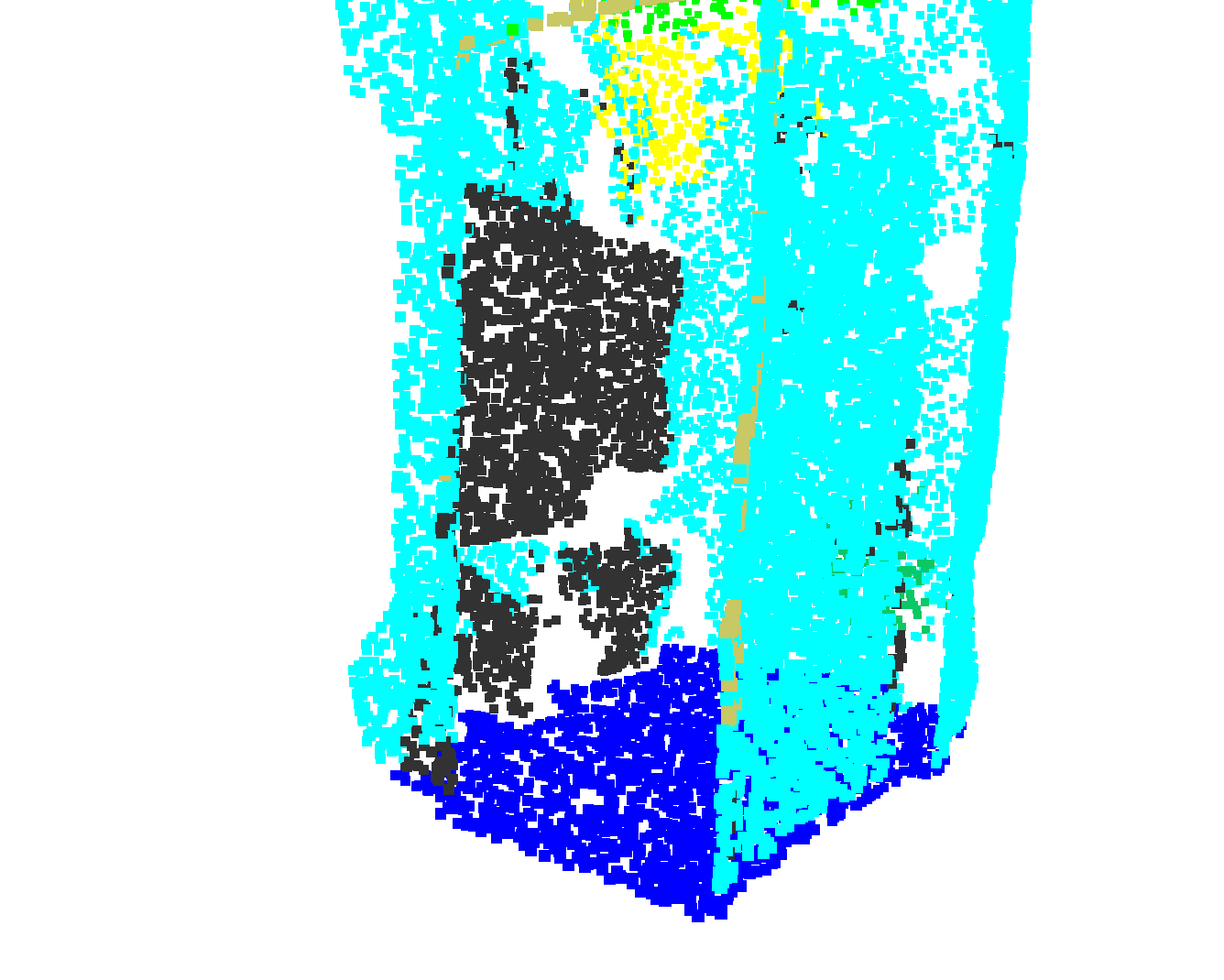} \hspace{8mm}%
\includegraphics[width=0.15\linewidth, trim=200 0 350 200, clip]{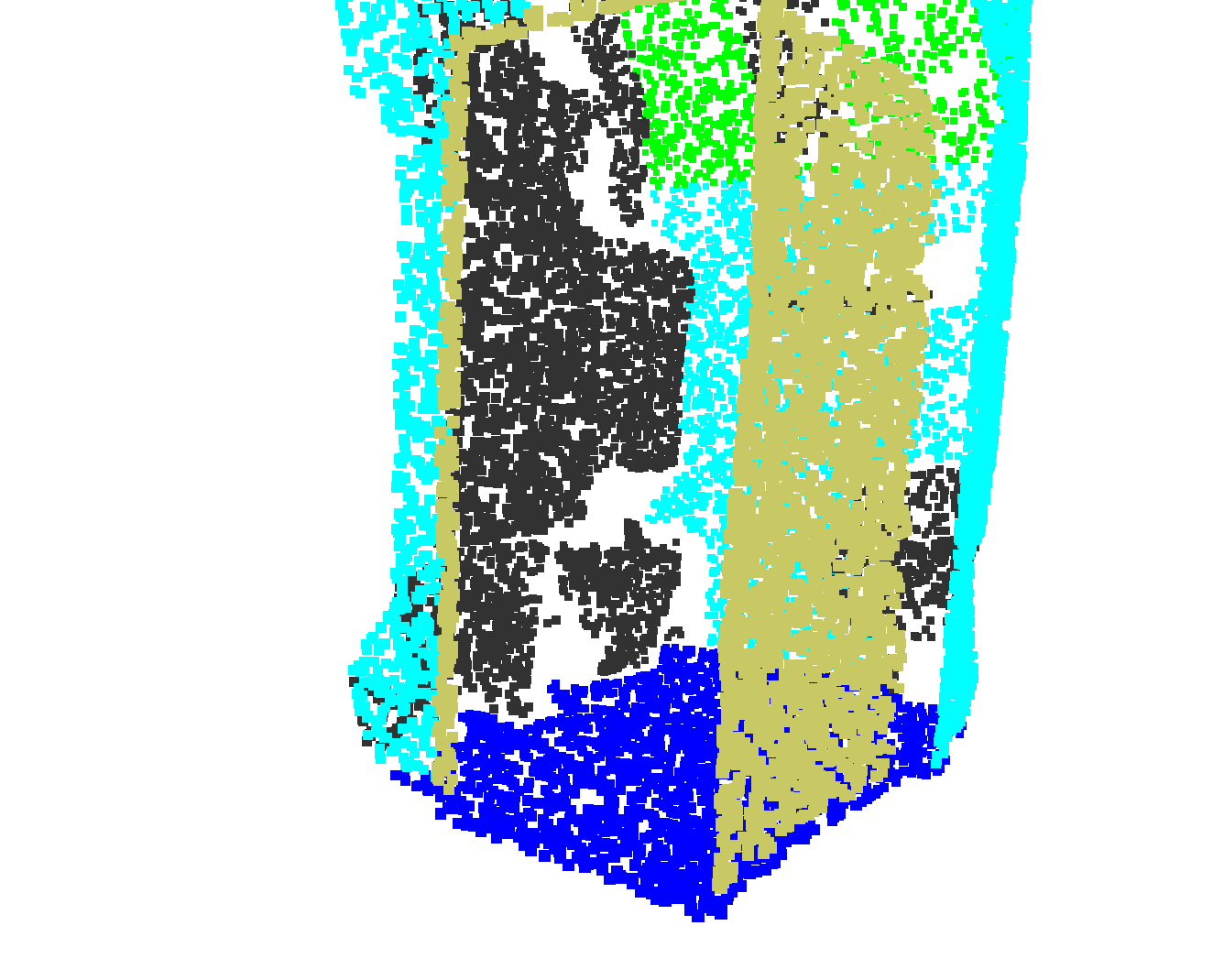} \hspace{8mm}%

\begin{minipage}{0.196\textwidth}
\caption*{\textbf{\small XYZ-RGB Input}}\label{fig:Scene}
\end{minipage}
\begin{minipage}{0.196\textwidth}
\caption*{\textbf{\small PointNet\cite{pointnet} }}\label{fig:PN}
\end{minipage}
\begin{minipage}{0.196\textwidth}
\caption*{\textbf{\small Ours, G-RCU}}\label{fig:GRC}
\end{minipage}
\begin{minipage}{0.196\textwidth}
\caption*{\textbf{\small Ours, MS-CU(2)}}\label{fig:MSCU}
\end{minipage}
\begin{minipage}{0.196\textwidth}
\caption*{\small \textbf{Ground Truth}}\label{fig:GT}
\end{minipage}
\vspace{-20px}
\caption{\textbf{Indoor qualitative results.} Dataset: S3DIS \cite{s3dis} with XYZ-RGB input features. From left to right: input point cloud, baseline method PointNet, our results using the G-RCU model (see \reffig{trans_model}), our results using the MS-CU(2) model (see \reffig{scale_model}), ground truth semantic labels. Our models produce more consistent and less noisy labels.}
	\label{fig:quali_results_s3dis}
\end{figure*}

We present qualitative results of our models applied to indoor scenarios in \reffig{quali_results_s3dis} and outdoor results in \reffig{quali_results_vkitti} along with a short discussion.
Additionally, we applied our pre-trained geometry-only model (vKITTI) to real-world laser data. The results are shown in \reffig{quali_results_laser} and \reffig{quali_results_3drms}.

\begin{table*}
\begin{center}
\begin{small}
\begin{tabular}[t]{lcccccccccccccc}
\toprule
 S3DIS Dataset \cite{s3dis} & mean IoU &
\rotatebox{70}{Ceiling}&%
\rotatebox{70}{Floor}&%
\rotatebox{70}{Wall}&%
\rotatebox{70}{Beam}&%
\rotatebox{70}{Column}&%
\rotatebox{70}{Window}&%
\rotatebox{70}{Door}&%
\rotatebox{70}{Table}&%
\rotatebox{70}{Chair}&%
\rotatebox{70}{Sofa}&%
\rotatebox{70}{Bookcase}&%
\rotatebox{70}{Board}&%
\rotatebox{70}{Clutter}\\
\midrule
*PointNet \cite{pointnet} & 43.5 &
81.5 &
86.7 &
64.8 &
29.4 &
16.3 &
39.1 &
48.1 &
52.5 &
42.5 &
5.4 &
37.6 &
30.4 &
31.4 \\

*MS & 44.4 &
82.2 &
86.9 &
64.2 &
33.8 &
22.8 &
43.3 &
52.0 &
51.0 &
38.6 &
9.2 &
36.1 &
23.6 &
33.7 \\

*MS + RCU & 45.5 &
83.6 &
86.9 &
\textbf{67.5} &
\textbf{40.5} &
17.1 &
37.0 &
48.8 &
\textbf{53.9} &
42.3 &
6.8 &
\textbf{39.7} &
\textbf{32.8} &
34.2\\

*SS + CU(1) & 45.9 &
\textbf{88.6} &
92.6 &
66.3 &
36.2 &
23.6 &
47.1 &
51.2 &
50.2 &
36.9 &
\textbf{12.6} &
33.7 &
22.7 &
35.3 \\

*MS + CU(2) & \textbf{47.8} &
\textbf{88.6} &
\textbf{95.8} &
67.3 &
36.9 &
\textbf{24.9} &
\textbf{48.6} &
\textbf{52.3} &
51.9 &
\textbf{45.1} &
10.6 &
36.8 &
24.7 &
\textbf{37.5} \\

\midrule
\midrule

PointNet \cite{pointnet}  & 47.6 &
88.0 &
88.7 &
\textbf{69.3} &
42.4 &
23.1 &
47.5 &
\textbf{51.6} &
54.1 &
42.0 &
\textbf{9.6} &
38.2 &
29.4 &
35.2 \\

G + RCU & \textbf{49.7} &
\textbf{90.3} &
\textbf{92.1} &
67.9 &
\textbf{44.7} &
\textbf{24.2} &
\textbf{52.3} &
51.2 &
\textbf{58.1} &
\textbf{47.4} &
6.9 &
\textbf{39.0} &
\textbf{30.0} &
\textbf{41.9} \\

\bottomrule

\end{tabular}
\end{small}
\end{center}
\vspace{-15px}
\caption{\textbf{IoU per semantic class. S3DIS dataset with XYZ-RGB input features.}
We compare our models with different components against the original PointNet baseline.
By adding different components, we can see an improvement of mean IoU.
We obtain state-of-the-art results in mean IoU and all individual class IoU.
\textit{Entries marked with * use random sampling for input block selection instead of discrete positions on a regular grid.}}
\label{tab:s3dis_color_iou}
\end{table*}

\begin{table*}
\begin{center}
\begin{small}
\begin{tabular}[t]{lcccccccccccccc}
\toprule
S3DIS Dataset \cite{s3dis}
  &
 mean IoU &
\rotatebox{70}{Ceiling}&%
\rotatebox{70}{Floor}&%
\rotatebox{70}{Wall}&%
\rotatebox{70}{Beam}&%
\rotatebox{70}{Column}&%
\rotatebox{70}{Window}&%
\rotatebox{70}{Door}&%
\rotatebox{70}{Table}&%
\rotatebox{70}{Chair}&%
\rotatebox{70}{Sofa}&%
\rotatebox{70}{Bookcase}&%
\rotatebox{70}{Board}&%
\rotatebox{70}{Clutter}\\
\midrule

*PointNet \cite{pointnet}  &
40.0 &
84.0 &
87.2 &
57.9 &
37.0 &
19.6 &
\textbf{29.3} &
35.3 &
\textbf{51.6} &
42.4 &
11.6 &
26.4 &
12.5 &
25.5 \\

*MS + CU(2) & \textbf{43.0} &
\textbf{86.5} &
\textbf{94.9} &
\textbf{58.8} &
\textbf{37.7} &
\textbf{25.6} &
28.8 &
\textbf{36.7} &
47.2 &
\textbf{46.1} &
\textbf{18.7} &
\textbf{30.0} &
\textbf{16.8} &
\textbf{31.2} \\

\midrule
vKITTI Dataset \cite{vkitti}
 &
 mean IoU &
\rotatebox{75}{Terrain}&%
\rotatebox{75}{Tree}&%
\rotatebox{75}{Vegetation}&%
\rotatebox{75}{Building}&%
\rotatebox{75}{Road}&%
\rotatebox{75}{GuardRail}&%
\rotatebox{75}{TrafficSign}&%
\rotatebox{75}{TrafficLight}&%
\rotatebox{75}{Pole}&%
\rotatebox{75}{Misc}&%
\rotatebox{75}{Truck}&%
\rotatebox{75}{Car}&%
\rotatebox{75}{Van} \\
\hline
*PointNet \cite{pointnet} & 17.9 &
32.9 &
76.4 &
11.9 &
17.7 &
49.9 &
3.6 &
2.8 &
3.7 &
3.5 &
0.7 &
1.5 &
25.1 &
3.4 \\

*MS + CU(2) & \textbf{26.4} &
\textbf{38.9} &
\textbf{87.1} &
\textbf{14.6} &
\textbf{44.0} &
\textbf{58.4} &
\textbf{12.4} &
\textbf{9.4} &
\textbf{10.6} &
\textbf{5.3} &
\textbf{2.2} &
\textbf{3.6} &
\textbf{43.0} &
\textbf{13.3}\\
\bottomrule
\end{tabular}
\end{small}
\end{center}
\vspace{-15px}
\caption{\textbf{IoU per semantic class. S3DIS and vKITTI datasets both with XYZ input features (no color)}. Our methods not only outperform PointNet consistently on two datasets, the improvements in mean IoU are also more considerable when no color is available. This suggests that our network architectures are able to learn improved geometric features and are more robust to varying point densities as they occur in the outdoor vKITTI dataset. 
}
\label{tab:s3dis_iou}
\end{table*}

\begin{table}[h]
\begin{center}
\begin{tabular}[t]{ l c c c }
\toprule
                             & mean & overall   & avg. class \\
                             & IoU  & accuracy  & accuracy \\
  \midrule
    \multicolumn{4}{ c}{S3DIS Dataset \cite{s3dis} -- no RGB} \\
  \midrule
  *PointNet \cite{pointnet}  & 40.0 & 72.1 & 52.9 \\
  *MS + CU(2)                      & \textbf{43.0} & \textbf{75.4} & \textbf{55.2}\\
   \midrule
    \multicolumn{4}{ c }{vKITTI Dataset \cite{vkitti} -- no RGB} \\
  \midrule
  *PointNet \cite{pointnet}  & 17.9 & 63.3 & 29.9 \\
  *MS + CU(2)                      & \textbf{26.4} & \textbf{73.2} & \textbf{40.9} \\%
  \bottomrule

\end{tabular}
\end{center}
\vspace{-15px}
\caption{\small \textbf{S3DIS and vKITTI datasets with only XYZ input features, without RGB.} We show improved results on indoor (S3DIS) and outdoor (vKITTI) datasets. Our presented mechanisms are even more important when no color is available.}
\label{tab:s3dis}
\end{table}

\begin{table}[h]
\begin{center}
\begin{tabular}{l c c c}
\toprule
S3DIS Dataset \cite{s3dis}  & mean  & overall  & avg. class \\
XYZ-RGB  & IoU   & accuracy & accuracy \\
\midrule
*PointNet \cite{pointnet}  & 43.5 & 75.0 & 55.5 \\
*MS          & 44.4 & 75.5 & 57.6 \\
*MS + RCU     & 45.5 & 77.2 & 57.2 \\
*SS + CU(1) & 45.9 & 77.8 & 57.7 \\
*MS + CU(2) & \textbf{47.8} & \textbf{79.2} & \textbf{59.7} \\
\midrule
\midrule

 PointNet \cite{pointnet}  &    47.6 & 78.5   &  66.2\\
 G + RCU                     &  \textbf{49.7} & \textbf{81.1} & \textbf{66.4}\\
\bottomrule
\end{tabular}
\end{center}
\vspace{-15px}
\caption{\small \textbf{S3DIS Dataset with XYZ-RGB input features.} Comparison of different context expansion techniques on input- and output-level (see Sections \ref{sec:input_level_context}--\ref{sec:consolidations}).
MS: Multi-Scale, SS: Single-Scale, G: Grid, CU: Consolidation Unit, RCU: Recurrent Consolidation Unit.
 \textit{Entries marked with * use random sampling for input block selection instead of discrete positions on a regular grid.}}
\label{tab:s3dis_color}
\end{table}

\section{Conclusion}
In this work, we investigated the question how to incorporate spatial context into a neural network architecture for 3D semantic segmentation. Building upon PointNet, we proposed two extension (Input-level context and Output-level context) which we successfully applied onto indoor and outdoor datasets.
Still, numerous other combinations remain possible. The full exploration of the design space is left for future work.

\newpage
\clearpage
\clearpage
\hspace{-13px}
\begin{minipage}[t]{1\textwidth}
	\begingroup
	\parfillskip=-5pt
  \begin{minipage}[t]{0.99\textwidth}
\begin{centering}
\colorbox{Terrain}{\strut Terrain}
\colorbox{Tree}{\strut \textcolor{white}{Tree}}
\colorbox{Vegetation}{\strut Vegetation}
\colorbox{Building}{\strut Building}
\colorbox{Road}{\strut \textcolor{white}{Road}}
\colorbox{Car}{\strut Car}
\colorbox{Truck}{\strut \textcolor{white}{Truck}}
\colorbox{Van}{\strut Van}
\colorbox{GuardRail}{\strut GuardRail}
\colorbox{TrafficSign}{\strut TrafficSign}
\colorbox{TrafficLight}{\strut TrafficLight}
\colorbox{Pole}{\strut \textcolor{white}{Pole}}
\colorbox{Misc}{\strut Misc}

\includegraphics[width=0.3\linewidth]{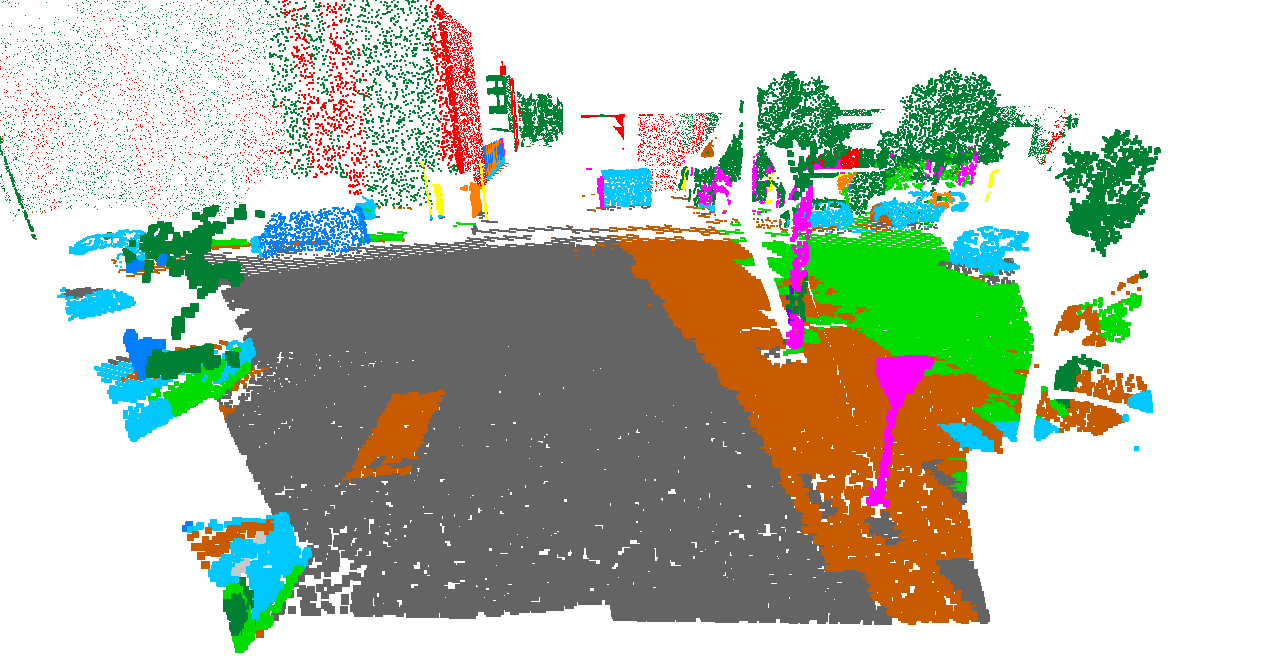}% pointnet
\includegraphics[width=0.3\linewidth]{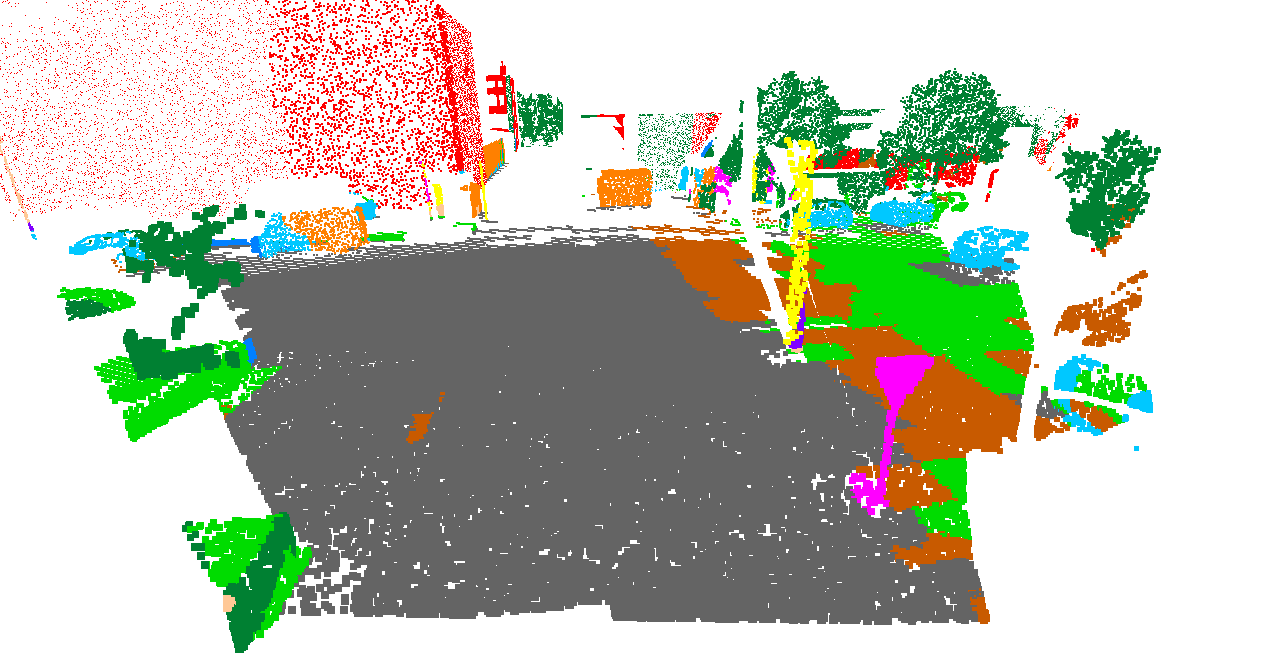}% ours
\includegraphics[width=0.3\linewidth]{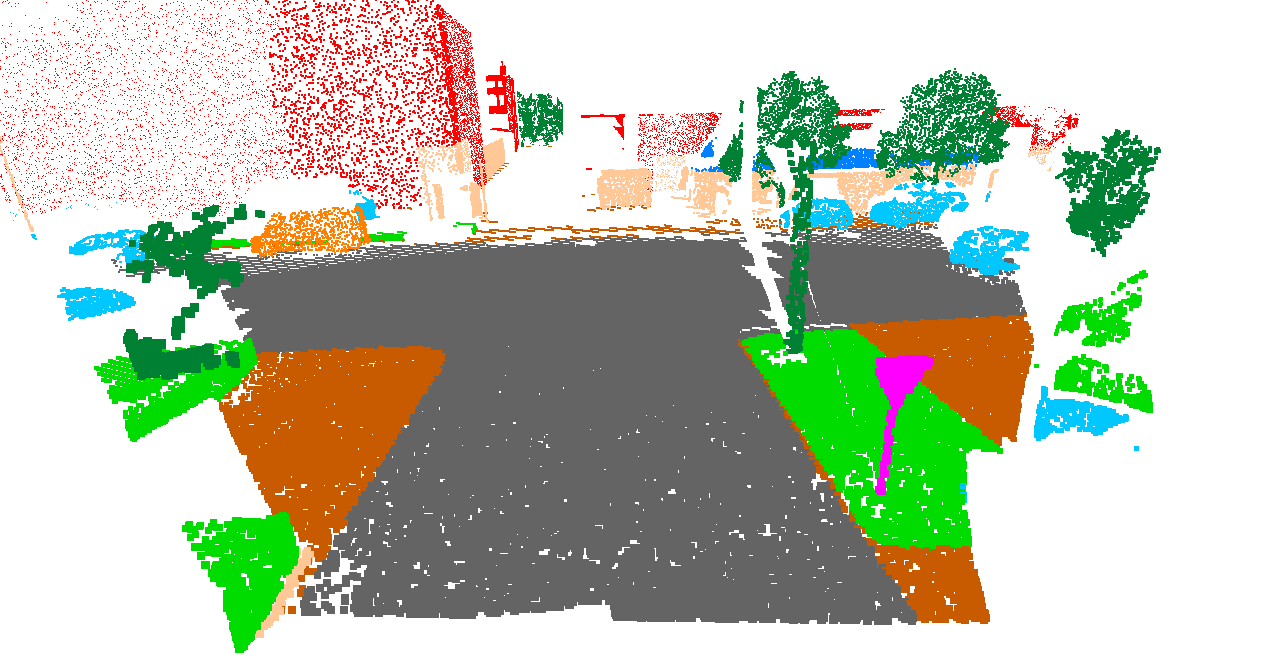}% groundtruth

\includegraphics[width=0.3\linewidth]{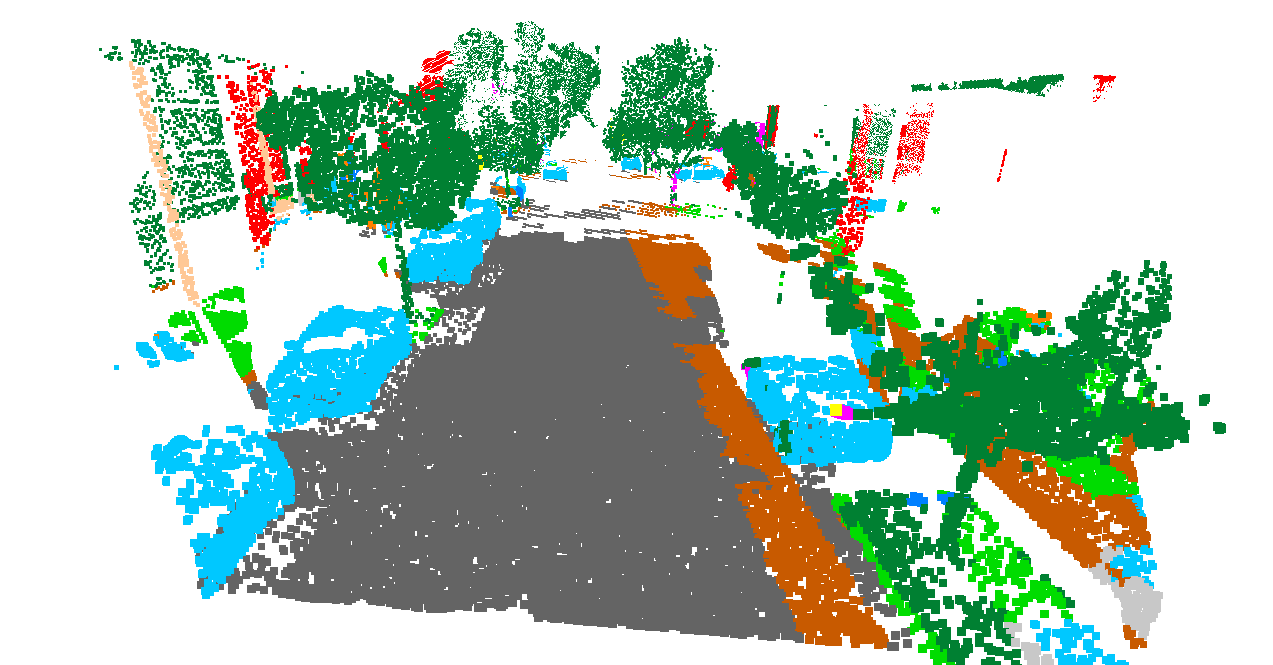}%
\includegraphics[width=0.3\linewidth]{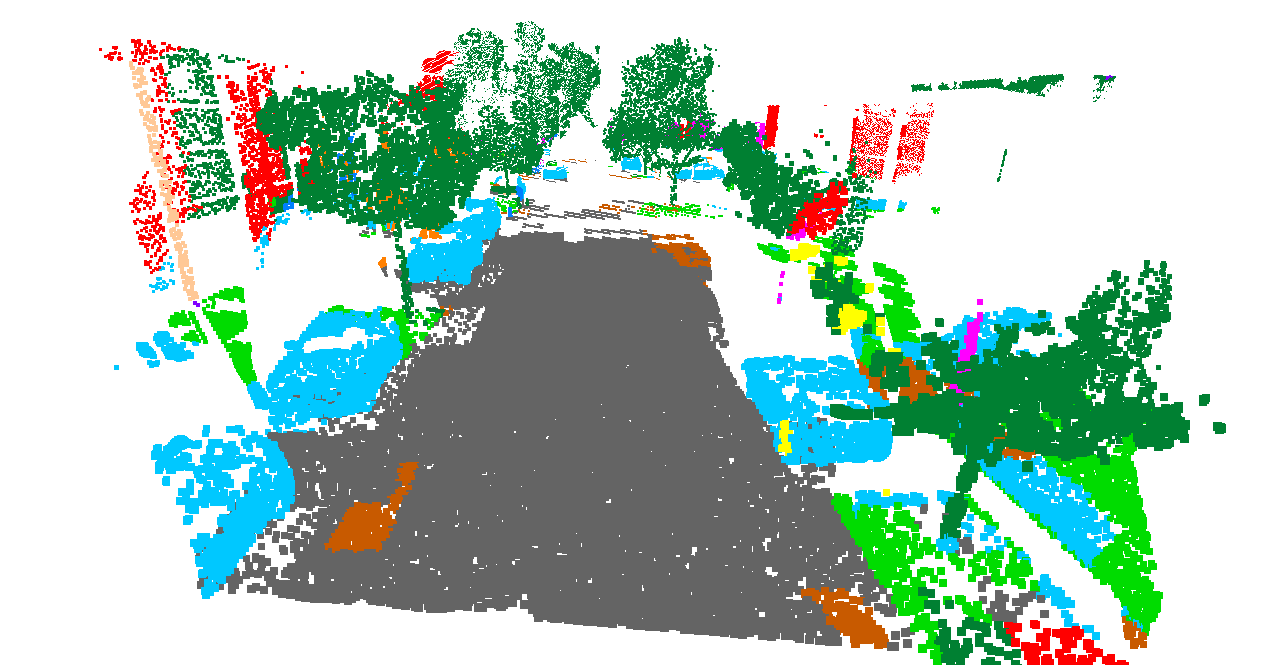}%
\includegraphics[width=0.3\linewidth]{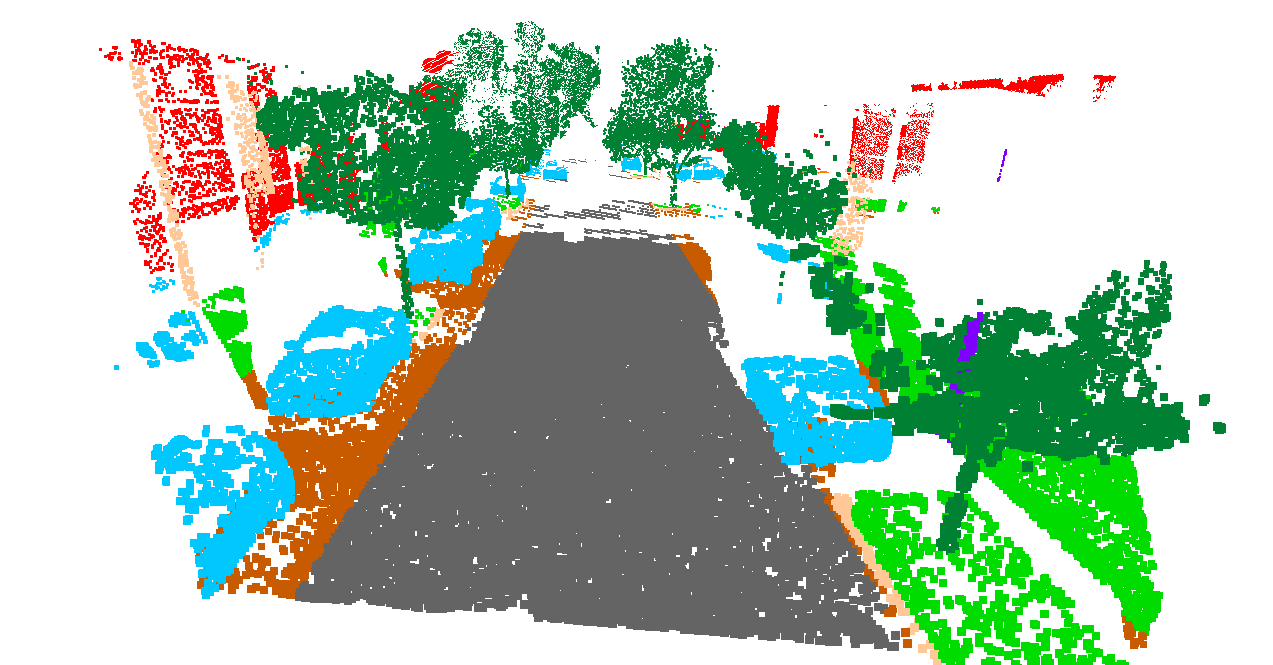}%

\includegraphics[width=0.3\linewidth]{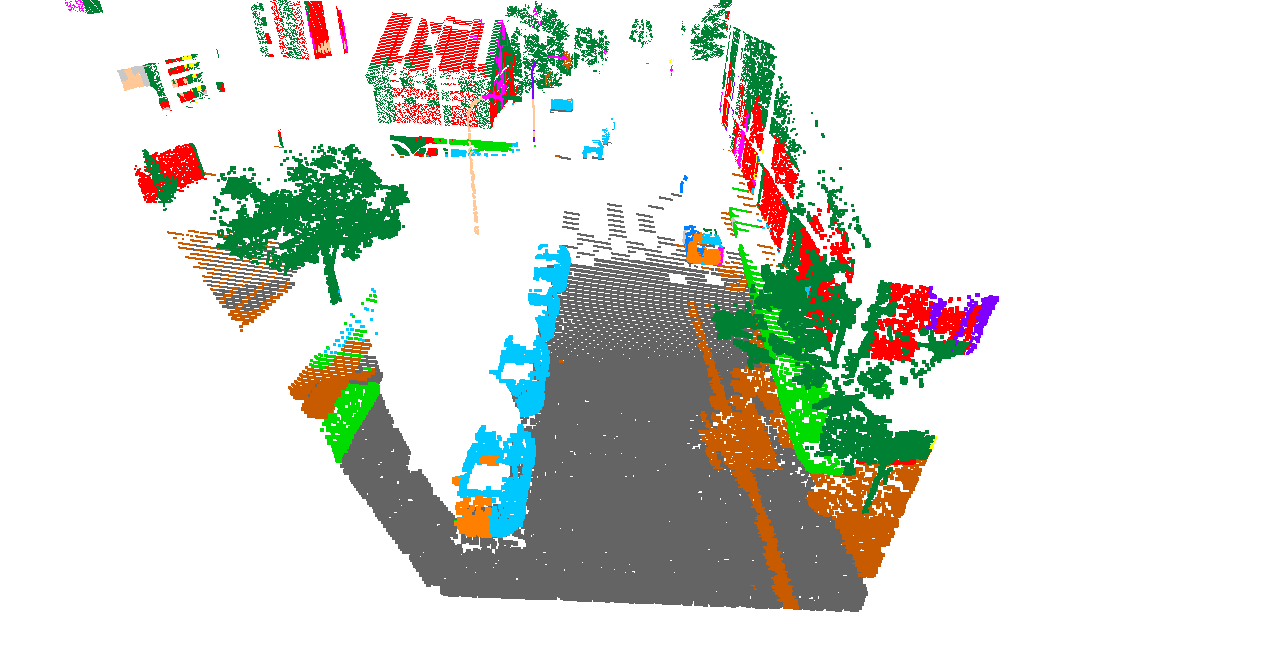}%
\includegraphics[width=0.3\linewidth]{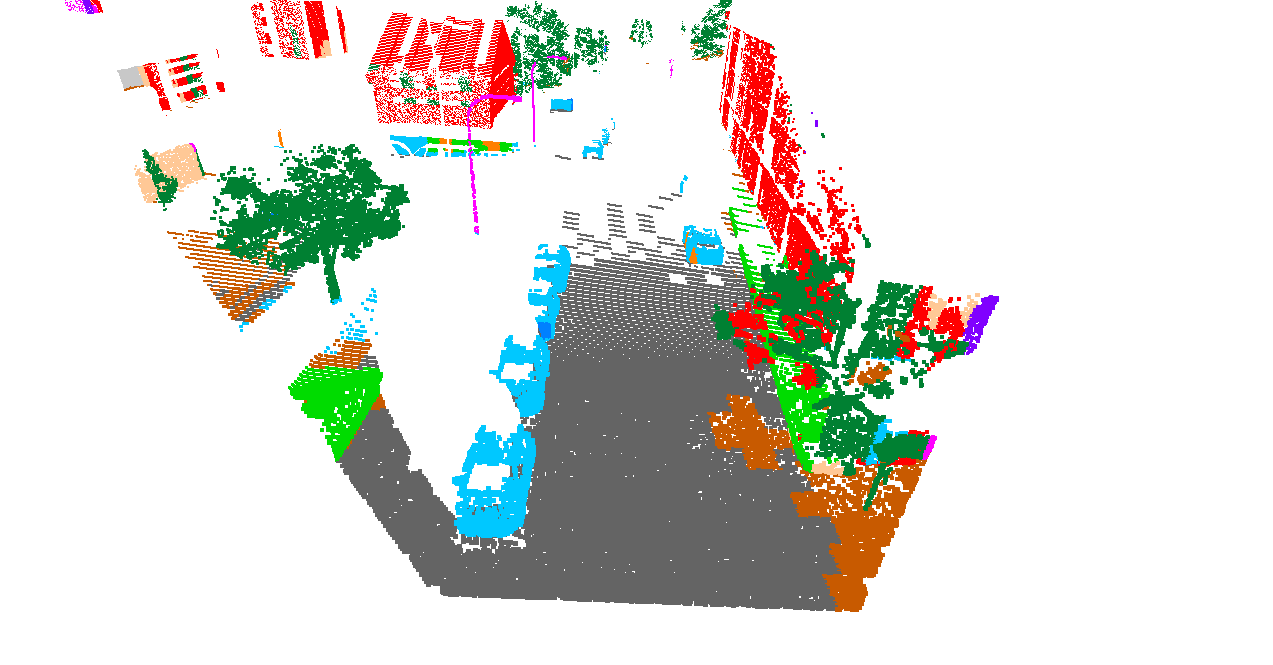}%
\includegraphics[width=0.3\linewidth]{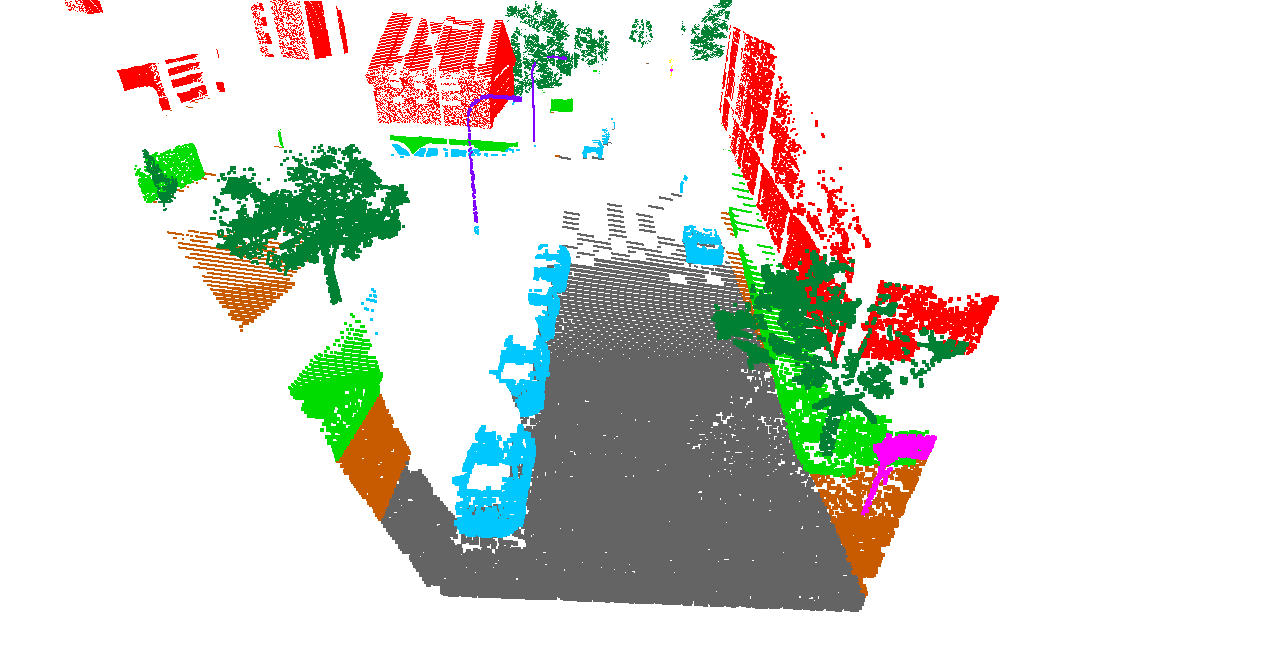}\\%
\end{centering}
\begin{minipage}{0.33\textwidth}
\centering{\textbf{\small PointNet \cite{pointnet}}}\label{fig:Scene}
\end{minipage}
\begin{minipage}{0.33\textwidth}
\centering{\textbf{\small Ours, MS-CU(2)}}\label{fig:PN}
\end{minipage}
\begin{minipage}{0.33\textwidth}
\centering{\textbf{\small Ground Truth}}\label{fig:GRC}
\end{minipage}
\\ \\
{{Figure\,8: \textbf{Outdoor qualitative results.} Dataset: Virtual KITTI \cite{vkitti}. Results were obtained using only XYZ coordinates as input, no color information was used. Left: baseline method PointNet. Center: our results using the MS-CU model as illustrated in \reffig{scale_model}. Right: ground truth semantic labels. The outputs of our method are less fragmented (cars, houses) and finer structures like street lights and poles are recognized better.}}
	\vspace{10mm}
	\end{minipage}%
	\\
	\begin{minipage}[t]{0.475\textwidth}
\vspace{0mm}
\begin{footnotesize}
\textbf{
\colorbox{Tree_}{\makebox(25,8){\centering \textcolor{white}{Tree}}}
\colorbox{Grass_}{\makebox(25,8){\centering \textcolor{white}{Grass}}}
\colorbox{Topiary_}{\makebox(30,8){\centering \textcolor{white}{Topiary}}}
\colorbox{Ground_}{\makebox(35,8){\centering \textcolor{white}{Ground}}}
\colorbox{Obstacle_}{\makebox(35,8){\centering \textcolor{white}{Obstacle}}}
\colorbox{Unknown_}{\makebox(35,8){\centering \textcolor{white}{Unknown}}}
}
\vspace{3px}

\includegraphics[width=0.49\linewidth, trim=100 120 100 90, clip]{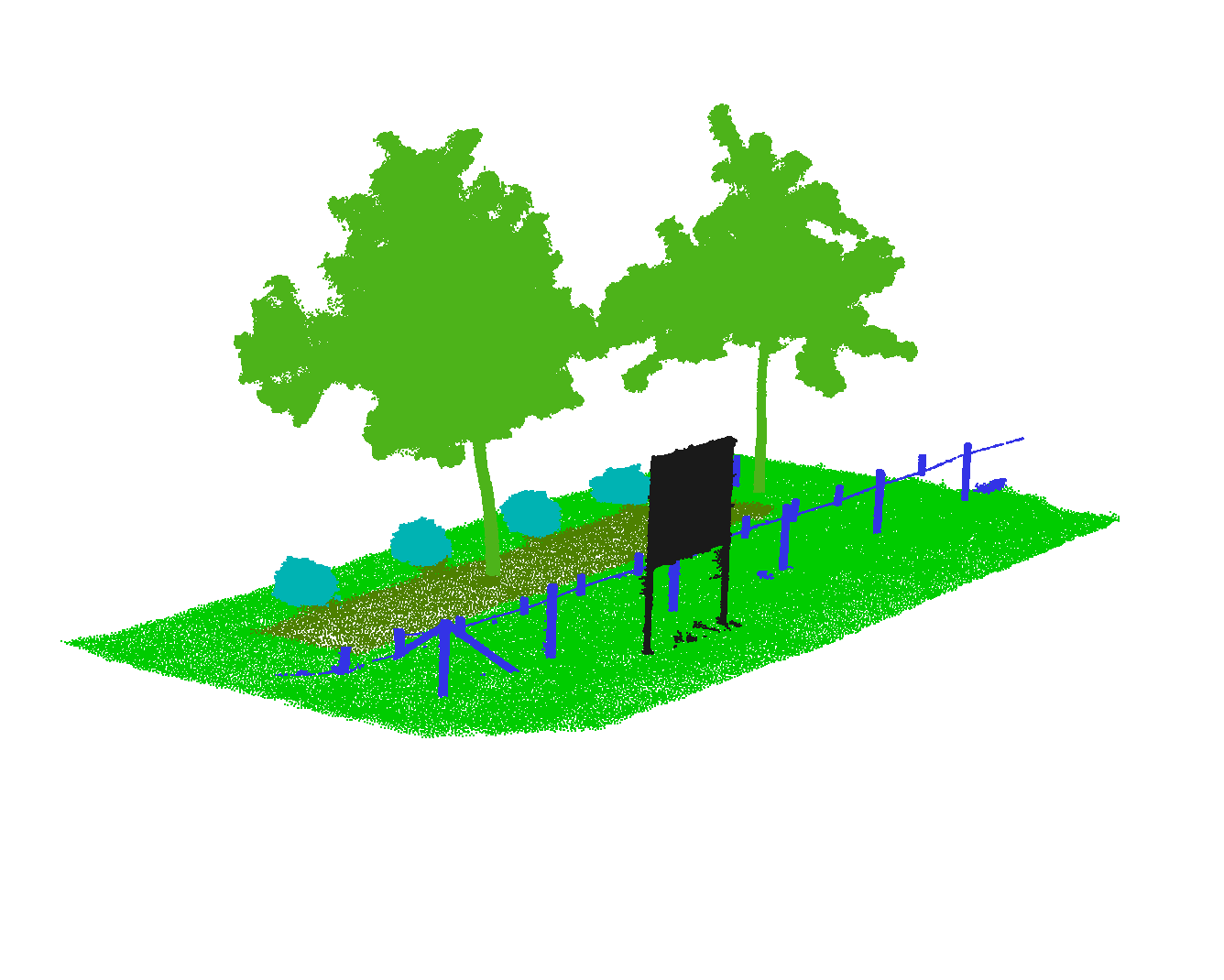}%
\includegraphics[width=0.49\linewidth, trim=100 120 100 90, clip]{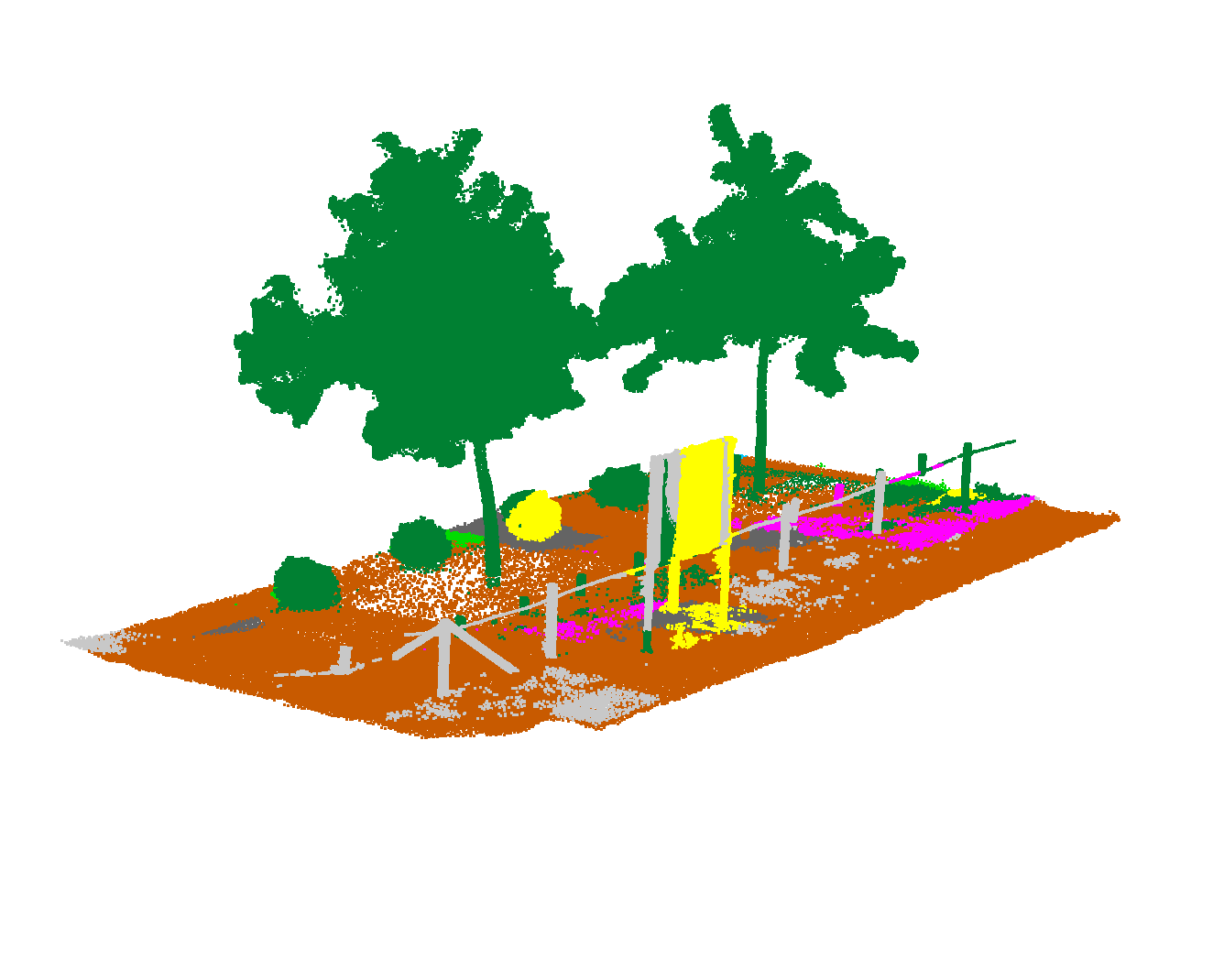}

\begin{minipage}{0.45\textwidth}%
\centering
\vspace{-5px}
{\textbf{Ground truth}\\ Labels: top}\label{fig:3drms_gt}%
\end{minipage}%
\hfill
\begin{minipage}{0.45\textwidth}%
\centering
\vspace{-5px}
{\textbf{Our prediction}\\ Labels: below}\label{fig:3drms_pred}%
\end{minipage}%
\vspace{1px}%
\\
\textbf{
\colorbox{Terrain}{\strut \textcolor{white}{Terrain}}\hspace{1px}%
\colorbox{Tree}{\strut \textcolor{white}{Tree}}\hspace{1px}%
\colorbox{Vegetation}{\strut \textcolor{white}{Vegetation}}\hspace{1px}%
\colorbox{GuardRail}{\strut {GuardRail}}\hspace{1px}%
\colorbox{TrafficSign}{\strut \textcolor{white}{TrafficSign}}\hspace{1px}%
\colorbox{TrafficLight}{\strut {TrafficLight}}%
}
\end{footnotesize}
\\ \\
{Figure\,9:\,\textbf{Qualitative results on 3DRMS'17 Challenge.}
We trained our model on vKITTI point clouds without color and applied it to the 3DRMS laser data.
Training and test datasets do not have the same semantic labels. Despite that, common classes like trees are successfully segmented and plausible ones are given otherwise (e.g. terrain instead of grass, guardrail instead of obstacle).}
	
	\end{minipage}%
	\hfill
	\begin{minipage}[t]{0.475\textwidth}%
\section{Acknowledgment}
We are grateful to our colleagues for providing valuable feedback on the paper and having fruitful discussions, especially with Umer Rafi and Paul Voigtlaender.
This work was supported by the ERC Starting Grant project CV-SUPER (ERC-2012-StG-307432).
	\end{minipage}%
	\hspace{7px}
	\par\endgroup	
\end{minipage}

\clearpage{\thispagestyle{empty}\cleardoublepage}
{\small
\bibliographystyle{ieee}
\bibliography{egbib}
}

\end{document}